\RequirePackage{fix-cm}
\documentclass{article}
\ifdefined\pdfminorversion\pdfminorversion=7\fi
\usepackage[T1]{fontenc}
\usepackage{times,natbib}
\setcitestyle{authoryear,round,citesep={;},aysep={,},yysep={;}}
\renewcommand{\footnoterule}{\kern-3pt\hrule width 12pc\kern2.6pt}
\renewcommand{\topfraction}{.95}
\renewcommand{\textfraction}{.08}

\makeatletter
\renewcommand{\normalsize}{\@setfontsize\normalsize{10}{11}}
\renewcommand{\small}{\@setfontsize\small{9}{10}}
\renewcommand{\footnotesize}{\@setfontsize\footnotesize{9}{10}}
\renewcommand{\scriptsize}{\@setfontsize\scriptsize{7}{8}}
\renewcommand{\tiny}{\@setfontsize\tiny{6}{7}}
\renewcommand{\large}{\@setfontsize\large{12}{14}}
\renewcommand{\Large}{\@setfontsize\Large{14}{16}}
\renewcommand{\LARGE}{\@setfontsize\LARGE{17}{20}}
\renewcommand{\huge}{\@setfontsize\huge{20}{23}}
\renewcommand{\Huge}{\@setfontsize\Huge{25}{28}}
\normalsize
\renewcommand{\section}{\@startsection{section}{1}{\z@}{-2ex plus -.5ex minus -.2ex}{1.5ex plus .3ex minus .2ex}{\large\scshape\raggedright}}
\renewcommand{\subsection}{\@startsection{subsection}{2}{\z@}{-1.8ex plus -.5ex minus -.2ex}{.8ex plus .2ex}{\normalsize\scshape\raggedright}}
\renewcommand{\subsubsection}{\@startsection{subsubsection}{3}{\z@}{-1.5ex plus -.5ex minus -.2ex}{.5ex plus .2ex}{\normalsize\scshape\raggedright}}
\renewcommand{\paragraph}{\@startsection{paragraph}{4}{\z@}{1.5ex plus .5ex minus .2ex}{-1em}{\normalsize\bfseries}}
\renewcommand{\subparagraph}{\@startsection{subparagraph}{5}{\z@}{1.5ex plus .5ex minus .2ex}{-1em}{\normalsize\scshape}}
\renewcommand{\@maketitle}{\vbox{\hsize\textwidth\parindent\z@\parskip\z@\raggedright
  {\fontsize{19}{22}\selectfont\bfseries\@title\par}
  \vskip 10pt
  {\fontsize{10.5}{12.5}\selectfont\bfseries\@author\par}
  \vskip 7pt
  {\normalfont\fontsize{9}{11}\selectfont\AuthorAffiliations\par}
  \vskip 6pt}}
\renewenvironment{abstract}{\par\vspace{4pt}\begingroup\fontsize{10}{12}\selectfont
  \noindent\textbf{Abstract}\par\vspace{3pt}\noindent}{\par\endgroup\vspace{6pt}}
\makeatother
\usepackage{amsmath,amssymb,amsthm,mathtools}
\usepackage{graphicx,booktabs,tabularx,longtable,array,multirow}
\usepackage{epstopdf}
\usepackage{caption}
\usepackage[font=footnotesize]{subfig}
\usepackage{microtype,enumitem}
\usepackage{algorithm,algpseudocode}
\usepackage{placeins,needspace}
\usepackage{xcolor,tikz,colortbl}
\definecolor{OursRow}{HTML}{EAF3F8}
\usetikzlibrary{arrows.meta,positioning,calc,fit}
\definecolor{PaperTeal}{HTML}{326F80}
\definecolor{PaperInk}{HTML}{263F4A}
\definecolor{PaperLight}{HTML}{EEF4F6}
\definecolor{PaperOchre}{HTML}{AF8D56}
\definecolor{PaperLilac}{HTML}{EEE9F1}
\usepackage{hyperref}
\usepackage[capitalise,noabbrev]{cleveref}
\hypersetup{colorlinks=true,linkcolor=blue,citecolor=blue,urlcolor=blue,pdftitle={Drafting Agents, Not Just Tokens: Speculative Recursive Self-Improvement for Agentic Systems}}
\newtheorem{proposition}{Proposition}
\newtheorem{lemma}[proposition]{Lemma}
\theoremstyle{definition}
\newtheorem{definition}[proposition]{Definition}

\newcommand{\Draft}{\textsc{DRAFT}}
\newcommand{\Verify}{\textsc{VERIFY}}

\newcommand{\E}{\mathbb E}

\newcommand{\clip}{\operatorname{clip}}

\newcolumntype{Y}{>{\raggedright\arraybackslash}X}
\newcolumntype{P}[1]{>{\raggedright\arraybackslash}p{#1}}
\makeatletter
\providecommand{\theHALG@line}{}
\renewcommand{\theHALG@line}{\arabic{algorithm}.\arabic{ALG@line}}
\makeatother

\newcommand{\AlgRef}[1]{\hyperref[#1]{Algorithm~\ref*{#1}}}

\newcommand{\AlgAnchor}[1]{\raisebox{\baselineskip}[0pt][0pt]{\hypertarget{#1}{}}}

\floatstyle{plaintop}
\newfloat{authorrecord}{htbp}{loar}
\floatname{authorrecord}{Table}
\makeatletter
\let\c@authorrecord\c@table
\makeatother

\newcommand{\best}[1]{\ensuremath{\mathbf{#1}}}
\newcommand{\second}[1]{\ensuremath{#1}}
\arrayrulecolor{PaperInk!55}

\newcommand{\Accept}{\textsc{ACCEPT}}

\usepackage{etoolbox}
\definecolor{ContentsInk}{HTML}{000000}
\makeatletter
\newif\ifappendixcontents\appendixcontentsfalse
\newcommand{\enableappendixcontents}{\appendixcontentstrue}
\newcommand{\listofappendices}{%
 \section*{Contents for Appendix}
 \begingroup\color{black}\hypersetup{linkcolor=blue}\setcounter{tocdepth}{2}
 \@starttoc{apc}\endgroup}
\let\savedaddcontentsline\addcontentsline
\renewcommand{\addcontentsline}[3]{%
 \ifappendixcontents\ifstrequal{#1}{toc}{\savedaddcontentsline{apc}{#2}{#3}}{\savedaddcontentsline{#1}{#2}{#3}}%
 \else\savedaddcontentsline{#1}{#2}{#3}\fi}

\makeatother
\newcommand{\emethod}{\textbf{\emph{SpecRSI}}}

\hypersetup{pdfsubject={SpecRSI: two-level speculative recursive self-improvement, verification-aware tree search, and group verification optimization}}

\newcommand{\fullmethod}{\textit{\textbf{Spec}ulative \textbf{R}ecursive \textbf{S}elf-\textbf{I}mprovement}}

\DeclareRobustCommand{\keypoint}[1]{\tikz[baseline=(p.base)]{\node[circle,fill=black,text=white,inner sep=1.1pt,font=\sffamily\bfseries\scriptsize](p){#1};}}
\definecolor{FrontMatterBlue}{HTML}{EFF5FC}
\definecolor{FrontMatterAccent}{HTML}{4525A0}
\makeatletter
\newcommand{\FrontMatterCard}{%
  \begingroup
  \noindent\hspace*{-14pt}%
  \begin{tikzpicture}
  \node[fill=FrontMatterBlue,draw=none,rounded corners=10pt,
        inner xsep=14pt,inner ysep=16pt,outer sep=0pt] {%
    \begin{minipage}{\textwidth}
    \setlength{\parindent}{0pt}\setlength{\parskip}{0pt}
    \noindent\hfill\includegraphics[width=0.32\textwidth]{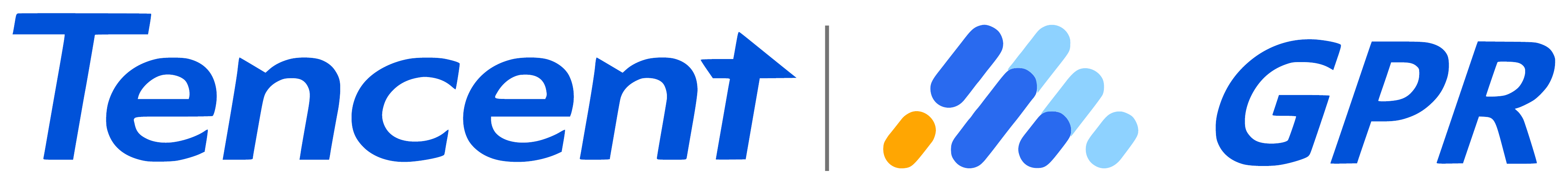}\par
    \vspace{7pt}
    {\raggedright\fontfamily{phv}\fontsize{19}{22}\selectfont\bfseries
     \@title\par}
    \vspace{10pt}
    {\raggedright\fontfamily{phv}\fontsize{10.5}{13}\selectfont\bfseries
     \@author\par}
    \vspace{7pt}
    {\raggedright\normalfont\fontsize{9.5}{11.5}\selectfont
     \AuthorAffiliations\par}
    \vspace{8pt}
    \begin{abstract}
Improving an agent through successive revisions requires balancing exploration and evaluation under a limited budget. A narrow search can miss useful sequences of edits, while fully evaluating every intermediate version often exhausts the budget before those sequences yield a verified improvement. Inspired by the draft--verify principle of speculative decoding, we propose \fullmethod{} (\emethod{}), a two-level framework coupling inner-loop agent search with outer-loop improver learning. \emethod{} drafts successors before their predecessors are fully evaluated, while adopting only fully evaluated, admissible versions. In the inner loop, Verification-Aware Tree Search (VATS) allocates the budget between further edits and outstanding checks using the predicted verified return of a continuation. In the outer loop, Group Verification Optimization (GVO) updates the improver from completed searches while keeping the real evaluator fixed. We evaluate \emethod{} on scientific optimization (including active queue management (AQM) for network infrastructure), coding, and general agent-system benchmarks. Under matched evaluation-call budgets, \emethod{} improves scientific terminal gain over Janus by 60.0\%--96.3\% and area under the budget curve by 32.1\%--91.6\%, while achieving higher means across all 50 scientific backbone--task pairs and outperforming MetaRSI-v1 by 1.30--2.70 percentage points (pp) in system score across twenty backbones. On instance-level comparisons, it further improves PyBaMM terminal gain by 142.6\% and SWE-Pro success by 6.0 pp at 7.0\% and 9.2\% lower token cost, respectively. These results demonstrate the efficacy of the proposed approach.
\end{abstract}

\begingroup
\fontsize{9}{11}\selectfont\setlength{\parskip}{3pt}\raggedright\urlstyle{same}
\noindent\textbf{Contact:} \href{mailto:pkushinnxu@tencent.com}{\nolinkurl{pkushinnxu@tencent.com}},
\href{mailto:neoxzhang@tencent.com}{\nolinkurl{neoxzhang@tencent.com}},
\href{mailto:zlin@pku.edu.cn}{\nolinkurl{zlin@pku.edu.cn}}.\par
\noindent\textbf{Code:} \url{https://github.com/SpecRSI-GPR/SpecRSI}.\par
\endgroup

    \end{minipage}};
  \end{tikzpicture}\hspace*{-14pt}\par
  \endgroup
  \vspace{10pt}}
\makeatother

\providecommand{\Draft}{\textsc{DRAFT}}
\providecommand{\Verify}{\textsc{VERIFY}}

\providecommand{\E}{\mathbb E}

\providecommand{\clip}{\operatorname{clip}}

\providecommand{\Full}{\mathcal C_{\mathrm{full}}}

\providecommand{\Apply}{\operatorname{Apply}}
\providecommand{\Freeze}{\operatorname{Freeze}}

\providecommand{\Generate}{\operatorname{Generate}}
\providecommand{\Acquire}{\operatorname{Acquire}}

\providecommand{\Debit}{\operatorname{Debit}}

\makeatletter
\@ifundefined{proposition}{}{}
\@ifundefined{lemma}{}{}
\@ifundefined{corollary}{}{}
\@ifundefined{definition}{}{}
\makeatother

\providecommand{\Remaining}{\operatorname{Remaining}}
\providecommand{\Jobs}{\operatorname{Jobs}}

\title{Drafting Agents, Not Just Tokens:\\\textcolor{FrontMatterAccent}{Speculative} Recursive Self-Improvement\\for Agentic Systems}
\author{
\hypersetup{urlcolor=black}
\mbox{\href{mailto:pkushinnxu@tencent.com}{Xin Xu}\textsuperscript{1,4,\ensuremath{\dagger}}},
\mbox{\href{mailto:ckpassenger@buaa.edu.cn}{Ke Cheng}\textsuperscript{1,6,\ensuremath{\dagger}}},
\mbox{\href{mailto:haohaow@zju.edu.cn}{Hao Wang}\textsuperscript{2,\ensuremath{\dagger}}},
\mbox{\href{mailto:yarayrhuang@tencent.com}{Yangru Huang}\textsuperscript{1,\ensuremath{\dagger}}},
\mbox{\href{mailto:xiaoyanghao22@mails.ucas.ac.cn}{Yanghao Xiao}\textsuperscript{3,\ensuremath{\dagger}}},
\mbox{\href{mailto:12032042@zju.edu.cn}{Zhichao Chen}\textsuperscript{4,\ensuremath{\dagger}}},
\mbox{\href{mailto:zszhong@link.cuhk.edu.hk}{Zhisheng Zhong}\textsuperscript{5}},
\mbox{\href{mailto:reubenliu@tencent.com}{Xun Liu}\textsuperscript{1}},
\mbox{\href{mailto:sincereli@tencent.com}{Yi Li}\textsuperscript{1}},
\mbox{\href{mailto:kathieqli@tencent.com}{Qian Li}\textsuperscript{1}},
\mbox{\href{mailto:haibinlv@tencent.com}{Haibin Lv}\textsuperscript{1}},
\mbox{\href{mailto:neoxzhang@tencent.com}{Jun Zhang}\textsuperscript{1,*}},
\mbox{\href{mailto:zeus@tencent.com}{Jie Jiang}\textsuperscript{1}},
\mbox{\href{mailto:zlin@pku.edu.cn}{Zhouchen Lin}\textsuperscript{4,*}}
}
\newcommand{\AuthorAffiliations}{
\textsuperscript{1}Tencent AMS, Beijing, China;\quad
\textsuperscript{2}Xiaohongshu;\quad
\textsuperscript{3}University of Chinese Academy of Sciences;\quad
\textsuperscript{4}State Key Laboratory of General Artificial Intelligence, School of Intelligence Science and Technology, Peking University, Beijing, China;\quad
\textsuperscript{5}Tencent Hunyuan;\quad
\textsuperscript{6}CCSE lab, Beihang University.\par
\vspace{4pt}\textsuperscript{\ensuremath{\dagger}}Equal contribution.\quad\textsuperscript{*}Corresponding authors.\par
}
\hypersetup{pdfauthor={Xin Xu, Ke Cheng, Hao Wang, Yangru Huang, Yanghao Xiao, Zhichao Chen, Zhisheng Zhong, Xun Liu, Yi Li, Qian Li, Haibin Lv, Jun Zhang, Jie Jiang, Zhouchen Lin}}

\usepackage[skins,breakable]{tcolorbox}
\newenvironment{findingblock}{\par\begingroup}{\par\endgroup}

\newcommand{\Description}[1]{} 
\newcommand{\CaptionRoman}{\fontfamily{ptm}\selectfont\upshape\mdseries}
\DeclareCaptionFont{captionroman}{\CaptionRoman\upshape\mdseries\color{black}\hypersetup{allcolors=black}}
\begin{document}
\ifdefined\XeTeXversion\fi
\FrontMatterCard
\begingroup
\raggedbottom
\setlength{\parskip}{4pt}
\setlength{\abovedisplayskip}{6pt plus 1pt minus 2pt}
\setlength{\belowdisplayskip}{6pt plus 1pt minus 2pt}
\begin{figure}[t]
\centering
\includegraphics[width=1.0\columnwidth]{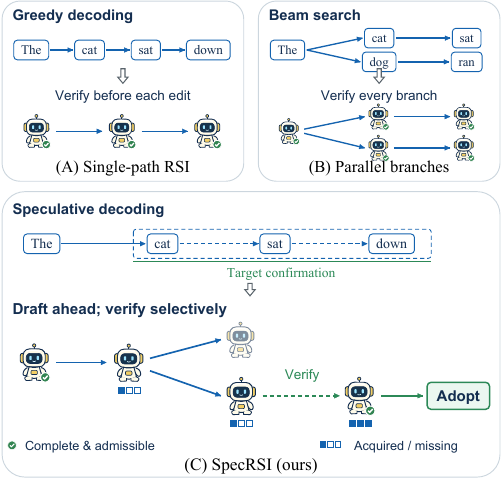}
\caption{From token decoding to agent self-improvement. Gray arrows mark structural analogies, while green dashed arrows indicate verification of the same agent version. SpecRSI drafts ahead, verifies selectively, and adopts only fully evaluated, admissible versions.}
\label{fig:intro}
\Description{Three panels map token decoding patterns to agent self-improvement through gray downward arrows: greedy decoding to single-path RSI, beam search to parallel branches, and speculative decoding to SpecRSI. Blue arrows generate successors. SpecRSI drafts from partially evaluated parents and selectively verifies a chosen agent version; adoption requires complete evidence and admissibility. Filled and empty squares indicate acquired and missing check outcomes.}
\end{figure}
\section{Introduction}
Agent systems can improve through successive revisions to their prompts, tools, and workflows, and recursive self-improvement (RSI) extends this goal to the mechanisms that generate subsequent improvements. Recent systems revise agent programs \citep{dgm2025,hgm2025}, modify the meta-agent that proposes revisions \citep{hyperagents2026}, adapt improvement operators and their scheduling in MetaRSI \citep{metarsi2026}, or co-evolve target programs and proxy evaluators to allocate expensive real evaluations in Janus \citep{janus2026}. Across these approaches, proposing changes and establishing their benefits both consume time and computation. This allocation problem is central to Web infrastructure and agentic systems: in active queue management (AQM), revising a controller to manage queue buildup and delay requires assessing traffic delivery, delay, loss, and fairness across diverse scenarios, consuming substantial simulator resources~\citep{janus2026,almeida2024desired}. This raises a central question: \textbf{how should an agent allocate its limited budget between exploring new revisions and verifying their benefits?} Useful revisions often support later improvements, yet identifying them requires expensive evaluation; the challenge is turning exploration into verified progress before resources run out (Figure~\ref{fig:intro}).

Two limiting strategies expose this trade-off. Following one fully evaluated lineage gives every decision fresh feedback, but narrows the alternatives and places an evaluation barrier between dependent edits---for instance, when a coding-agent revision adds a file--test map that a successor uses for focused intermediate retesting while retaining the final full suite. Maintaining many lineages explores more broadly, but fully evaluating them all spends budget on branches that are never used. Extra workers cut elapsed time without making evaluation free, and under a fixed budget a wider search may not leave enough computation to finish the most useful branch. While prior systems partially relax this barrier by separating expansion from evaluation or validating in parallel \citep{hgm2025,flashevolve2026}, we focus on drafting successors ahead of their parents' full evaluation and jointly selecting further revisions and bundles of outstanding checks under the remaining budget.

Allowing successors to be drafted ahead changes where the budget should be spent next. A candidate that looks strong may still need several checks before it can be adopted, whereas a child that repairs an unhelpful parent may need only one remaining check. Evaluating a version solely by its apparent quality misses the cost of making it usable. Under a hard budget, what matters is a \emph{continuation}---a sequence of further edits and checks from the current state---and the verified improvement it can still deliver with the remaining budget. Because this return must be estimated \emph{before} paying for the checks that reveal it, scoring must shift from static candidate quality to the marginal verified return of further editing and checking.

Extending speculation from tokens to agent revisions therefore changes the search space, the verification process, and the learning objective. Agent versions are revisable programs whose successors may repair or build on a parent that is not yet admissible. Verification requires paid, complementary external checks with heterogeneous costs and dependencies; several may be jointly necessary for adoption, so a single target-model pass cannot supply this evidence. Progress must consequently be measured by realized verified gains under resource limits, accounting for all incurred search costs, including failed and unfinished work.

\noindent\textbf{Contributions.} We introduce \emethod{}, a framework for budgeted speculative agent improvement. \keypoint{1}~\textbf{Two-Level Speculative Self-Improvement.} We decouple speculative exploration from verified adoption: inner searches extend incompletely evaluated versions but adopt only fully evaluated, admissible ones, while the improver remains frozen within each search and learns from complete-search feedback between rounds. \keypoint{2}~\textbf{Verification-Aware Tree Search (VATS).} VATS allocates the remaining budget across drafting and verification: the Solver preserves affordable completion bundles, including individually low-value checks; the Scorer predicts verified gains attainable through further edits and checks within the remaining budget; and the Policy selects the next draft or verification plan. \keypoint{3}~\textbf{Group Verification Optimization (GVO).} GVO assigns credit across independent complete searches from the same start, using verified gains from searches under matched resource caps to provide delayed credit to sampled edits and work choices for the Drafter and Policy, while training the Scorer on realized suffix returns and paid same-state comparisons.
\section{Preliminaries and Related Work}\label{sec:related}
\subsection{Speculative Decoding}\label{sec:spec-decoding}
Speculative decoding uses a cheap draft model $q$ to propose tokens ahead of a target model $p$ \citep{leviathan2023}. Given prefix $x_{<i}$, let $q_i=q(\cdot\mid x_{<i})$ and $p_i=p(\cdot\mid x_{<i})$:
\begin{equation}
 x_i\sim q_i,\qquad \alpha_i=\min\!\left\{1,\frac{p_i(x_i)}{q_i(x_i)}\right\},
 \label{eq:spec-decoding}
\end{equation}
where $i$ indexes the token position, $x_i$ is the sampled draft token, and $\alpha_i$ is its acceptance probability. Evaluating a draft block in parallel, the target model retains the accepted prefix; at the first rejection, it samples a replacement token from the normalized positive part of $p_i-q_i$ and discards the remaining drafts, preserving the target distribution. \emethod{} brings this draft--verify separation to agent revisions (Figure~\ref{fig:intro}), where acceptance requires complete admissible evidence instead of token-distribution equivalence (extended related work is in Appendix~\ref{app:extended-related}).

\subsection{Recursive Self-Improvement}\label{sec:rsi-prelim}
In recursive self-improvement (RSI), we distinguish an agent version $c$---comprising executable code and persistent prompts, tools, and workflows---from an improver $\Theta$ that proposes and selects revisions; updating $\Theta$ changes how subsequent rounds generate improvements. Given retained versions $\mathcal V_t$ and observed history $H_t$, a proposal--evaluation step within round $k$ is
\begin{equation}
\begin{gathered}
 \widetilde c_t\sim\operatorname{Propose}_{\Theta_k}(\cdot\mid\mathcal V_t,H_t),
 \quad z_t=f_{\rm real}(\widetilde c_t),\\
 (\mathcal V_{t+1},H_{t+1})=\operatorname{Update}(\mathcal V_t,H_t,\widetilde c_t,z_t),
\end{gathered}\label{eq:rsi-basic}
\end{equation}
where $t$ indexes steps under fixed $\Theta_k$, $\widetilde c_t$ is the proposed version, and $z_t$ is its evaluation outcome. A full real evaluation returns $f_{\rm real}(c)=(J(c),v(c))$, with utility $J(c)$ (higher is better) and admissibility indicator $v(c)\in\{0,1\}$. Based on $z_t$, $\operatorname{Update}$ revises $\mathcal V_t$ and $H_t$; retained stepping stones need not immediately outperform the incumbent.
\section{Method}\label{sec:method}
\subsection{Motivation}\label{sec:motivation}
\begin{figure}[!htbp]
\centering
\captionsetup[subfloat]{position=bottom,justification=centering,farskip=1pt,nearskip=1pt}
\includegraphics[width=1.0\linewidth]{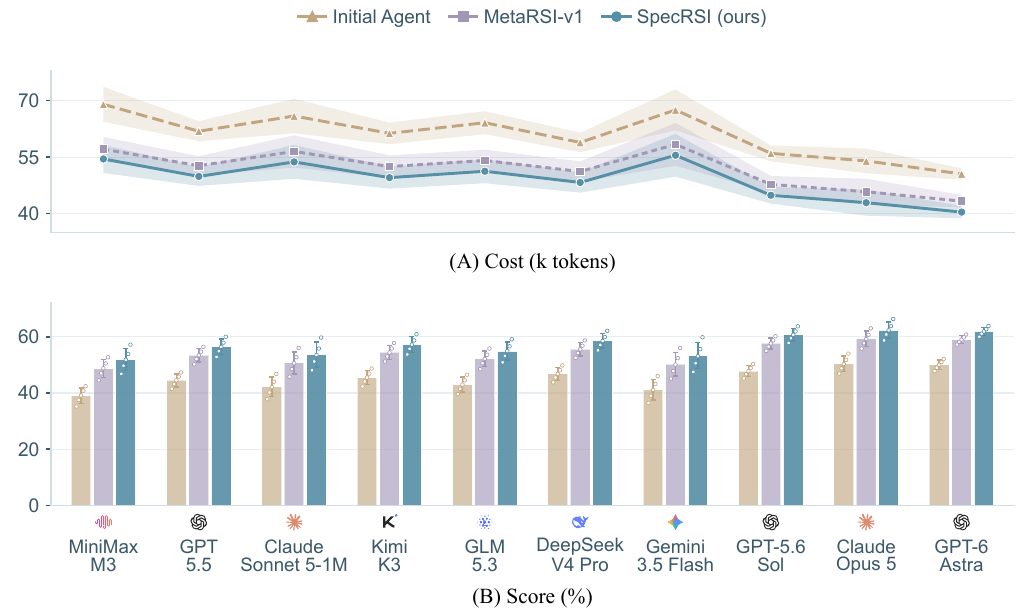}

\caption{Performance across ten backbones (five-seed means, with bands and whiskers showing $\pm$ SD). (A) Cost (k tokens). (B) Score (\%). Lines connect model categories, and logos denote model families.}
\label{fig:motivation}
\Description{Cost lines and score bars compare Initial Agent, MetaRSI-v1, and SpecRSI across ten backbones using the updated five-seed records. Cost bands and score whiskers show SD. SpecRSI has lower mean cost and higher mean score than MetaRSI-v1 throughout.}
\end{figure}

Accurate benefit evaluation is central to guiding RSI toward better agents under a limited search budget. If benefit evaluation is inaccurate, search either overlooks useful multi-step edits or wastes resources on candidates that never become admissible. Effective allocation must therefore evaluate the verified improvement that continued work can deliver.

Estimating this benefit is non-trivial because a candidate's current quality does not determine the value of continuing from it, due to two challenges. \keypoint{1}~\textbf{Benefit evaluation should reflect downstream verified improvement.} Intermediate revisions may serve as stepping stones to later gains, but realizing those gains requires subsequent edits and real checks. \keypoint{2}~\textbf{Benefit evaluation should account for completion costs.} Complementary checks, planning, and failed attempts all draw from the same finite resources, so even promising gains may be too costly to verify. Together, these challenges induce a performance--cost trade-off: a worthwhile continuation must offer sufficient verified improvement while remaining affordable to complete. Figure~\ref{fig:motivation} illustrates this trade-off across ten backbones, and Section~\ref{sec:resource-tradeoffs} analyzes broader lifecycle expenditure.

\begin{figure}[!htb]
\centering
\captionsetup[subfloat]{position=bottom,justification=centering,farskip=1pt,nearskip=1pt}
\makebox[\linewidth][c]{%
\begin{minipage}{\linewidth}
\centering
\subfloat[Two-level Speculative RSI Loop]{\begin{minipage}[b]{1.0\linewidth}\centering\includegraphics[width=\linewidth]{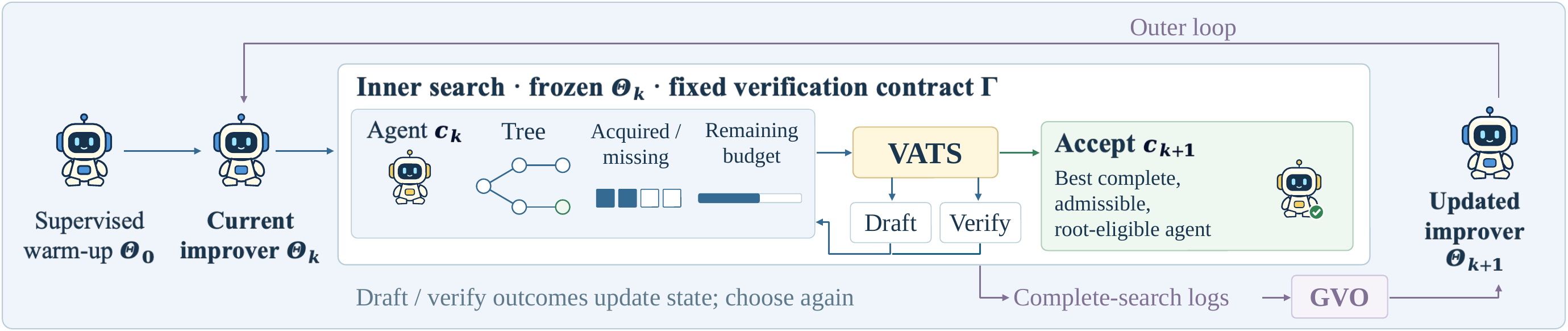}\end{minipage}}\par
\subfloat[Verification-Aware Tree Search (VATS)]{\begin{minipage}[b]{1.0\linewidth}\centering\includegraphics[width=\linewidth]{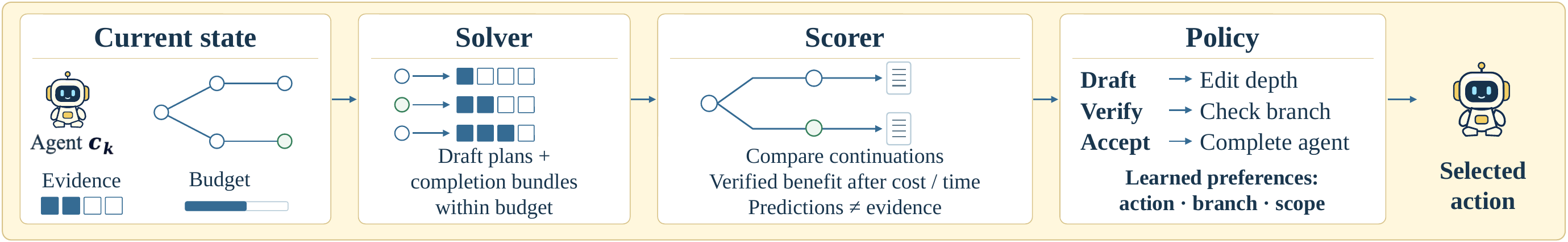}\end{minipage}}\par
\subfloat[Group Verification Optimization (GVO)]{\begin{minipage}[b]{1.0\linewidth}\centering\includegraphics[width=\linewidth]{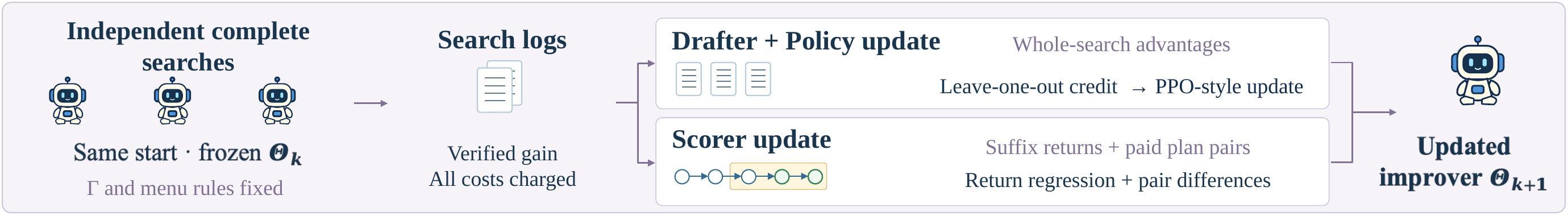}\end{minipage}}
\end{minipage}}
\caption{Two-level Speculative RSI Loop. Within each search, VATS allocates drafts and checks under a frozen improver before Accept returns the best complete admissible version; across searches, GVO updates the improver under a fixed contract $\Gamma$.}
\label{fig:overview}
\Description{Three diagrams show the frozen-improver search loop, Solver--Scorer--Policy allocation in VATS, and complete-search learning with GVO.}
\end{figure}

\subsection{Two-level Speculative RSI Loop}\label{sec:plans}\label{sec:framework}
SpecRSI brings the draft--verify separation of speculative decoding to the RSI loop, allowing successor agent versions to be drafted before their parents are fully evaluated while allocating real checks selectively. Unlike token acceptance in Equation~\eqref{eq:spec-decoding}, adopting an agent requires complete admissible evidence. Specifically, we refine Equation~\eqref{eq:rsi-basic} by organizing retained versions $\mathcal V_t$ as nodes in a search tree $\mathcal T=(\mathcal C,\mathcal A)$ anchored at a complete admissible root $r$, where $\mathcal C$ stores immutable agent versions and $\mathcal A$ records executed edit edges, and by decomposing $f_{\rm real}$ into paid checks. Each version freezes its prompts, tools, workflows, and permitted persistent state (including memory), with model-parameter edits allowed only under D/H/M permissions. An uneditable verification contract $\Gamma$ specifies a finite required-item set $\mathcal I$, execution and admission rules, and constraints; its score $J_\Gamma$ and admissibility condition instantiate the utility $J$ and indicator $v$ from Section~\ref{sec:rsi-prelim}. At any step, state $S=(\mathcal T,\mathcal E,H,B,N,n_{\mathrm{wrk}})$ tracks admitted evidence $\mathcal E$, history $H$, remaining expense budget $B\ge0$, remaining attempts $N\ge0$, and worker count $n_{\mathrm{wrk}}\ge1$. In token-accounted runs, $B$ and expense charges are measured in k tokens under the registered cap (while end-to-end totals also include overhead outside that cap), and elapsed time is tracked separately. For version $c$ and item $i$, $m_{ci}\in\{0,1\}$ indicates whether a valid outcome is present, regardless of whether it passed. The complete admissible set and its best observed gain are
\begin{equation}
\begin{gathered}
 \Full(S)=\{c\in\mathcal C:m_{ci}=1\ \forall i\in\mathcal I,\ c\text{ admissible}\},\\
 F(S)=\max_{c\in\Full(S)}[J_\Gamma(c)-J_\Gamma(r)],
\end{gathered}\label{eq:complete}
\end{equation}
where admissibility is judged under $\Gamma$ and $F(S)$ is the observed adoption-ready gain rather than a learned estimate; because root $r$ is eligible, search-time $F(S)\ge0$, while held-out gain is evaluated separately. The fixed contract $\Gamma$ and execution interface $\Acquire_\Gamma$ constitute the \emph{real evaluator}: $\Gamma$ defines valid evidence, $\Verify$ invokes $\Acquire_\Gamma$ to run selected real checks, and complete admissible evidence determines $J_\Gamma(c)$ and $\Accept$ eligibility. By contrast, the learned Scorer $\widehat Q_\phi(S,\omega)$ for proposed work $\omega$ serves as the \emph{score evaluator}, predicting continuation gain with optional expense and elapsed-time penalties (Equation~\eqref{eq:Q}) to guide budget allocation without supplying evidence, whereas $\Draft$ generates successor edits without assigning evaluation scores.

At outer round $k$, a complete admissible agent $c_k$ serves as root $r$. SpecRSI parameterizes the improver as $\Theta_k=(\theta_k,\psi_k,\phi_k,\nu)$, comprising Drafter parameters $\theta_k$, work-selection Policy parameters $\psi_k$, continuation-value Scorer parameters $\phi_k$, and fixed menu settings $\nu$. During search collection, all four components are frozen, with scorer snapshot $\bar\phi_k=\phi_k$. After $G\ge2$ independent inner searches (counting distinct search runs rather than shared-tree leaves), GVO updates the learned parameters:
\begin{equation}
\begin{gathered}
\text{\bfseries Inner loop:}\enspace 
(c_{k+1}^{(g)},\mathcal D_k^{(g)})
=\operatorname{Search}_{\Theta_k}(c_k),\quad g=1,\ldots,G,\\[2pt]
\text{\bfseries Outer loop:}\enspace 
\mathcal D_k=\bigcup\nolimits_{g=1}^G\mathcal D_k^{(g)},\quad
\Theta_{k+1}=\operatorname{GVO}(\Theta_k,\mathcal D_k),
\end{gathered}\label{eq:rsi-loop}
\end{equation}
where $g$ indexes the $G$ searches, each $\operatorname{Search}$ returns an admissible successor $c_{k+1}^{(g)}$ and trajectory log $\mathcal D_k^{(g)}$, and GVO updates $(\theta,\psi,\phi)$ from their union $\mathcal D_k$ while keeping $\nu$, $\Gamma$, and the outer update rule fixed (Figure~\ref{fig:overview}; see \AlgRef{alg:main} and Algs.~\ref{alg:prepare}--\ref{alg:epoch} in Appendix~\ref{app:methods}). Across outer rounds, \emph{Reset} returns the task root to $c_0$, whereas \emph{Carry} retains a complete admissible successor chosen by the registered inter-round selector; both choices operate independently of within-search acceptance and GVO updates, and neither resets the learned improver.

Whereas Janus co-evolves programs with proxy evaluators \citep{janus2026} and MetaRSI learns operator schedules \citep{metarsi2026}, SpecRSI learns to allocate budget between drafting and verification from complete-search verified returns (Appendix~\ref{app:extended-related}).
\subsection{Inner-loop Search with VATS}\label{sec:scorer}\label{sec:scoring}
Within each inner search, Verification-Aware Tree Search (VATS) iteratively selects bounded work $\omega$ under expense and attempt caps: either $\Draft(P,h)$, where $P=(P_a)_{a=1}^{|P|}$ indexes existing parent versions and $h\ge1$ bounds edit depth, or $\Verify(\mathcal J)$, where $\mathcal J$ specifies verification jobs on existing versions. Either action may repeat until $\Accept$ terminates the search. Because complementary checks cannot be scheduled by token-confidence ranking alone, VATS coordinates three modules under the frozen improver: the Solver constructs feasible completion bundles, the Scorer refines verification scope, and the Policy selects the next action before execution updates the state for replanning (Appendix~\ref{app:methods}).

\subsubsection{Solver}
The Solver assembles completion bundles that fit within the remaining budget. For version $c$, let $U_c(S)=\{i\in\mathcal I:m_{ci}=0\}$ denote missing items and $K_c(S)=\{(c,i):i\in U_c(S)\}$ the corresponding outstanding outcomes. A relation reader $\mathcal R(S)=\Remaining(S)$ extracts masks, dependencies, shared setups, and source relations from existing records without running new checks or substituting forecasts for missing evidence (Appendix~\ref{app:contract}). To ensure that versions can reach full admissibility even when individual checks carry little standalone value, the Solver forms bundled options:
\begin{equation}
 \Omega_t=\mathcal D_{\rm draft,t}\cup\{\Verify(A):A\in\mathcal M_t\}\cup\{\Accept\},\label{eq:tree-frontier}
\end{equation}
where $\Omega_t$ is the feasible work menu at decision $t$, $\mathcal D_{\rm draft,t}$ contains retained draft plans (which may extend incomplete parents), and $\mathcal M_t$ contains feasible verification bundles $A$. Specifically, a mixed-integer linear program (MILP) selects bundles covering target $K_c$ sets and cross-branch unions subject to prerequisites, shared setups, conflicts, and reserved budgets. While feasibility guarantees neither passing check outcomes nor exhaustive tree coverage, original completion seeds remain eligible throughout subsequent refinement.

\subsubsection{Scorer}
The Scorer $\widehat Q_\phi(S,\omega)$ estimates the continuation value of state--work pairs under the frozen procedure $\pi$ (which encompasses edit generation and work selection under fixed menu rules), accounting for subsequent edits, checks, and branch switches:
\begin{equation}
 Q^\pi(S_t,\omega)=\E_\pi\!\bigl[F(S_{\rm end})-F(S_t)-\lambda C_{t:{\rm end}}-\mu L_{t:{\rm end}}\mid S_t,\omega\bigr],
 \label{eq:Q}
\end{equation}
where $S_t$ is the state after settling the current preparation cost from pre-preparation state $S_t^-$ (Alg.~\ref{alg:prepare}), and $S_{\rm end}$ is the terminal state. Subsequent expense $C_{t:{\rm end}}$ and elapsed time $L_{t:{\rm end}}$ include later preparation and failed work but exclude sunk preparation at step $t$. Fixed weights $\lambda,\mu\ge0$ convert expense and elapsed time into score-unit penalties ($\mu=0$ in cost-only settings; our reference profile sets $\lambda=\mu=0$, relying on resource inputs and hard caps rather than explicit return penalties). By definition, $Q^\pi(S,\Accept)=0$, and Scorer predictions never replace observed $F$ or admitted evidence.

Given a feasible bundle $A$, let $D_j$ denote job $j$ together with its prerequisites. The frozen scorer snapshot $\widehat Q_{\bar\phi}$ refines $A$ by evaluating the marginal gain of adding $D_j$:
\begin{equation}
 \Delta_{\bar\phi}(j\mid A,\widetilde S)=
 \widehat Q_{\bar\phi}(\widetilde S,\Verify(A\cup D_j))-
 \widehat Q_{\bar\phi}(\widetilde S,\Verify(A)),
 \label{eq:marginal}
\end{equation}
where both terms are evaluated at the same conservative state $\widetilde S$ that reserves the preparation cap. Comparing whole bundles captures cross-check interactions without summing isolated check scores, counting shared jobs and setups only once. Query and extension caps bound refinement overhead while preserving the eligibility of initial feasible seeds (Appendix~\ref{app:plans}).

\subsubsection{Policy}
Once preparation cost is settled, the Policy selects work from the nonempty menu $\Omega_t$ conditioned on observed relations $\mathcal R_t=\Remaining(S_t)$. With temperature-scaled logits $\zeta_t(v)=z_\psi(S_t,\mathcal R_t,v)/T_{\rm pol}$, the sampling distribution is
\begin{equation}
 \pi_\psi(\omega\mid S_t,\mathcal R_t,\Omega_t)
 =(1-\epsilon)\frac{\exp(\zeta_t(\omega))}
 {\sum_{v\in\Omega_t}\exp(\zeta_t(v))}+\frac{\epsilon}{|\Omega_t|},
 \label{eq:actor}
\end{equation}
where $v\in\Omega_t$, $T_{\rm pol}>0$ is the sampling temperature, and $\epsilon\in[0,1)$ is the uniform-exploration rate. Because preference logits $z_\psi$ are parameterized separately from value estimates $\widehat Q$ (with the $\Accept$ logit fixed to zero), the Policy need not greedily maximize $\widehat Q$. Search collection samples from $\pi_\psi$, whereas deployment selects the maximum-logit action (Appendix~\ref{app:learning}).

\subsubsection{Execution and Acceptance}\mbox{}\par\nobreak
\noindent\textbf{DRAFT.} Given draft plan $(P,h)$, the Drafter initializes $c_{a,0}=P_a$ for each parent $a=1,\ldots,|P|$ and sequentially generates edits up to depth $d=1,\ldots,h$ using only the context $H_{a,d}$ observed prior to that edit:
\begin{equation}
 e_{a,d}\sim\pi_\theta(\cdot\mid c_{a,d-1},H_{a,d}),\quad
 c_{a,d}=\Freeze(\Apply(c_{a,d-1},e_{a,d})),
 \label{eq:draft}
\end{equation}
where $e_{a,d}$ is sampled from $\pi_\theta$, $\Apply$ executes the edit, and $\Freeze$ stores the resulting immutable version $c_{a,d}$. The executor $\Generate_\theta(P,h,S)$ runs this bounded plan on actual artifacts: incomplete parents may be extended, provided each edit waits for its immediate parent. Any invalid edit is charged to the budget and terminates that path while preserving its valid prefix.

\noindent\textbf{VERIFY.} For requested jobs $\Jobs(\omega)$, $\Acquire_\Gamma$ either retrieves valid identical records or executes real checks. Our core configuration runs direct checks only, whereas checked reuse imports results after scoped equality checks via the shared backend used by compatible controls. The executor records outcomes and updates the state before VATS replans (Appendix~\ref{app:execution}).

\noindent\textbf{ACCEPT.} Admitted records must satisfy identity, scope, field, and provenance requirements, including for failed checks. Upon termination, VATS adopts
\begin{equation}\label{eq:accept}
\text{ACCEPT}(S)\in\operatorname*{arg\,max}_{c\in\Full(S)}J_\Gamma(c),
\end{equation}
returning the root $r$ if it attains the maximum, and otherwise the first maximizer under a fixed version order. Only the adopted version requires complete admissible evidence, and the reference rule introduces no hidden completion or adoption margin.

\subsection{Outer-loop Learning with GVO}\label{sec:training}
Outer-loop training begins with supervised warm-up and then alternates between frozen-improver search collection and Group Verification Optimization (GVO) updates. Completed trajectories provide two supervision signals derived strictly from observed outcomes under the frozen improver rather than updated Scorer predictions: realized suffix returns train the Scorer (when return or pair records are present), and whole-search advantages train the Drafter and Policy over sampled choices (with empty empirical loss terms contributing zero).

\subsubsection{Supervised Warm-up}
Prior to the first collection round, supervised warm-up initializes $\Theta_0$ by training Drafter $\theta$ to imitate executed legal edits, Policy $\psi$ to imitate legal plan choices, and Scorer $\phi$ to predict observed returns; unexecuted alternatives receive no labels. Throughout training, Solver feasibility rules and menu settings $\nu$ remain fixed, and neither Solver outputs nor environment outcomes carry policy likelihoods.

\subsubsection{Search Returns and Advantages}
In each round, we run $G\ge2$ independent searches from a shared pre-preparation state $S_0^-$ under frozen $\Theta_k$, matching the root agent, verification contract, budgets, and worker count. Each search $g$ terminates at state $S_{\rm end}^g$ without requiring every visited version to be fully evaluated, logging total expense $C_g$ and elapsed time $L_g$ (including initial preparation as well as failed or unfinished work):
\begin{equation}
 \begin{gathered}
 R_g=F(S_{\rm end}^g)-F(S_0^-)-\lambda C_g-\mu L_g,\\
 A_g=R_g-\frac1{G-1}\sum_{u=1,\,u\ne g}^G R_u,
 \end{gathered}\label{eq:reward}
\end{equation}
where $R_g$ is the net return of search $g$ and $A_g$ is its leave-one-out advantage across the group $\{1,\ldots,G\}$. Here $F(S)$ is the best observed complete admissible gain over root $r$ (Equation~\eqref{eq:complete}), while $\lambda C_g$ and $\mu L_g$ penalize total cost and time in score units; if a search yields no new complete improvement, its gain is zero while incurred costs are still charged. Comparing whole searches eliminates the need for a learned baseline and assigns delayed credit to intermediate edits through their verified descendants, even when those intermediate versions are not themselves admissible.

\subsubsection{Scorer Training}
The Scorer is updated by minimizing a combined regression and pairwise objective,
$\min_{\phi}\bigl[\mathcal L_Q(\phi)+\beta_{\rm pair}\mathcal L_{\rm pair}(\phi)\bigr]$ with weight $\beta_{\rm pair}\ge0$.
The regression loss $\mathcal L_Q$ fits realized, undiscounted suffix returns from executed plan starts:
\begin{equation}
 \begin{gathered}
 G_t=F(S_{\rm end})-F(S_t)-\lambda C_{t:{\rm end}}-\mu L_{t:{\rm end}},\\
 \mathcal L_Q(\phi)=\E_{(S_t,\omega_t,G_t)\sim\mathcal D_Q}
 \big[(\widehat Q_\phi(S_t,\omega_t)-G_t)^2\big],\label{eq:valuefit}
 \end{gathered}
\end{equation}
where $G_t$ is the realized net return from executing $\omega_t$ at paid state $S_t$, and $\mathcal D_Q$ collects the resulting state--work--return triples under the frozen procedure (including paid alternatives). Here $F(S_{\rm end})-F(S_t)$ is the remaining observed gain, and $C_{t:{\rm end}}, L_{t:{\rm end}}$ measure expense and elapsed time from step $t$ to termination; unlike $R_g$, $G_t$ excludes sunk preparation at step $t$ while charging subsequent preparation, so $\widehat Q_\phi$ estimates the conditional expectation $Q^\pi$ via mean squared error.

To sharpen relative comparisons between legal plans $a,b$ at the same saved paid state $S$, paid rollouts under the shared continuation policy estimate mean returns $\overline G_a,\overline G_b$ and their difference $\widehat\delta_{ab}=\overline G_a-\overline G_b$:
\begin{equation}
 \mathcal L_{\rm pair}(\phi)=\E_{\mathcal D_{\rm pair}}
 \bigl[(\widehat Q_\phi(S,a)-\widehat Q_\phi(S,b)-\widehat\delta_{ab})^2\bigr],
 \label{eq:pair}
\end{equation}
where $\mathcal D_{\rm pair}$ stores matched tuples $(S,a,b,\widehat\delta_{ab})$. While $\mathcal L_Q$ anchors absolute value levels (which pairwise differences alone cannot identify), $\mathcal L_{\rm pair}$ directly supervises relative plan rankings. Generating these pairs restores paid state $S$, forces a legal initial plan, and continues under the frozen improver; the resulting rollouts train only the Scorer, with all clones and rollouts charged to the budget and logged failures retained as valid data (whereas missing records are never imputed as zero).

\subsubsection{Drafter and Policy Training}\label{sec:actor-training}
The actor parameters $\vartheta=(\theta,\psi)$ are updated over sampled edit tokens and plan choices via
$\min_{\vartheta}\bigl[\mathcal L_{\rm GVO}(\vartheta)+\beta_{\rm KL}\mathcal L_{\rm KL}(\vartheta)\bigr]$,
where $\mathcal L_{\rm GVO}$ reinforces choices from searches with positive advantage $A_g$ and $\beta_{\rm KL}\mathcal L_{\rm KL}$ regularizes divergence from the fixed post-warm-up actor (Appendix~\ref{app:kl-regularization}). Using the PPO-style clipped objective \citep{ppo2017} for sampled choice $y_{g,j}$ under context $H_{g,j}$ and legal support $\mathcal Y_{g,j}$, we have
\begingroup\small
\thinmuskip=2mu\medmuskip=2mu\thickmuskip=3mu
\begin{equation}
\begin{gathered}
\rho_{g,j}(\vartheta)=
\frac{p_\vartheta(y_{g,j}\mid H_{g,j},\mathcal Y_{g,j})}
{p_{\vartheta_k}(y_{g,j}\mid H_{g,j},\mathcal Y_{g,j})},\\
\mathcal L_{\rm GVO}(\vartheta)=-\frac1G\!\sum_{g=1}^G\!\sum_{j=1}^{J_g}\!\min\bigl\{\rho_{g,j}(\vartheta)A_g,\clip(\rho_{g,j}(\vartheta),1-\epsilon_c,1+\epsilon_c)A_g\bigr\},
\end{gathered}\label{eq:gvo}
\end{equation}
\endgroup
where $j$ indexes the $J_g$ sampled choices in search $g$, $\vartheta_k=(\theta_k,\psi_k)$ denotes the frozen collection parameters, and $\rho_{g,j}$ is the probability ratio between current and collection policies (evaluating Drafter probabilities for edit tokens and Policy probabilities for plan choices, with $\clip$ radius $\epsilon_c\in(0,1)$ distinct from exploration rate $\epsilon$). Trajectory logs $\mathcal B_t$ record token contexts, legal supports, and behavior probabilities (which are empty for $\Verify$ steps) so that numerator and denominator are evaluated under identical contexts, supports, temperatures, masks, and exploration settings. During gradient computation, collection probabilities and advantages are held constant, invalid sampled edits retain their likelihood and cost, and each choice contributes once without search-length normalization. In the single exact-replay step of Alg.~\ref{alg:epoch}, $\rho_{g,j}(\vartheta_k)=1$, so clipping is inactive and does not bound the update.

Finally, GVO freezes the updated improver $\Theta_{k+1}$ for the next collection round; because actor updates shift the continuation distribution, subsequent rounds collect fresh searches to recalibrate the Scorer under the fixed contract $\Gamma$, with Drafter and Policy supervision based on complete-search returns (Appendix~\ref{app:learning}).
\section{Experimental Results}\label{sec:exp}\label{sec:experiments}
In this section, we conduct a comprehensive empirical evaluation to demonstrate the efficacy of SpecRSI, centered on the following research questions (RQs):

\begin{enumerate}[label=\arabic*., labelindent=0pt, leftmargin=*, labelsep=0.5em, nosep]

\item \textbf{RQ1:} \emph{Does SpecRSI produce better successor agents?}
Section~\ref{sec:overall-performance} compares SpecRSI against competitive baselines across tasks and backbones under matched evaluation settings.

\item \textbf{RQ2:} \emph{How do verification scope and planning in VATS affect search quality and cost?}
Section~\ref{sec:search-evidence} examines verification scope, draft-ahead, and planning, with additional scorer and evidence-input controls in Appendix~\ref{app:results-planning}.

\item \textbf{RQ3:} \emph{Does GVO learn a more effective improver?}
Section~\ref{sec:improver-rounds} compares frozen and GVO-updated improvers from identical starting agents and tracks cumulative gains over successive rounds.

\item \textbf{RQ4:} \emph{How do planning capacity and parallelism trade off gain, cost, and latency?}
Section~\ref{sec:resource-tradeoffs} analyzes candidate-plan limits, full-lifecycle token expenditure, and multi-worker execution latency.
\end{enumerate}

\subsection{Evaluation Setup}\label{sec:experimental-setup}
\noindent\textbf{Datasets.}
We evaluate five scientific tasks (PyBaMM~\citep{sulzer2021pybamm}, AQM~\citep{almeida2024desired}, FDTD~\citep{wu2024cwdm}, Reactor~\citep{braniff2024reactor}, and IRP~\citep{violi2024inventory})\footnote{\href{https://github.com/pybamm-team/PyBaMM}{PyBaMM}; \href{https://github.com/nsnam/ns-3-dev-git}{ns-3}. Our implementations: \href{https://github.com/SpecRSI-GPR/SpecRSI/blob/main/src/specrsi/science/fdtd.py}{FDTD}, \href{https://github.com/SpecRSI-GPR/SpecRSI/blob/main/src/specrsi/science/reactor.py}{Reactor}, \href{https://github.com/SpecRSI-GPR/SpecRSI/blob/main/src/specrsi/science/irp.py}{IRP}.}
and four agent-system benchmarks (Terminal-Bench 2.1~\citep{terminalbench2026}, SWE-Pro~\citep{swebenchpro2025}, GPQA-hard100~\citep{gpqa2024}, and AIME).\footnote{\href{https://www.tbench.ai/news/terminal-bench-2-1}{Terminal-Bench 2.1}; \href{https://github.com/scaleapi/SWE-bench_Pro-os}{SWE-Pro}; \href{https://github.com/idavidrein/gpqa}{GPQA}; \href{https://maa.org/maa-invitational-competitions/}{AIME}.}
We further evaluate specialized settings for code (SWE-Pro~\citep{swebenchpro2025}, HumanEval-Ext.~\citep{chen2021humaneval}, MBPP-Advanced~\citep{austin2021mbpp}, FixBench~\citep{li2025fixbenchrtl}),\footnote{\href{https://github.com/scaleapi/SWE-bench_Pro-os}{SWE-Pro}; \href{https://github.com/openai/human-eval}{HumanEval}; \href{https://github.com/google-research/google-research/blob/master/mbpp/README.md}{MBPP}; \href{https://huggingface.co/datasets/KSU-HW-SEC/Fixbench-RTL/tree/main}{Fixbench-RTL}.}
prompt (GPQA-hard100~\citep{gpqa2024}, MMLU-Pro~\citep{wang2024mmlupro}, GSM8K-Opt.~\citep{cobbe2021gsm8k}),\footnote{\href{https://github.com/idavidrein/gpqa}{GPQA-hard100}; \href{https://github.com/TIGER-AI-Lab/MMLU-Pro}{MMLU-Pro}; \href{https://github.com/openai/grade-school-math}{GSM8K-Opt}.}
and workflow optimization (PlanBench~\citep{valmeekam2023planbench}, Math~\citep{emas2026}, MBPP~\citep{austin2021mbpp}, Game24~\citep{tot2023}).\footnote{\href{https://github.com/karthikv792/LLMs-Planning}{PlanBench}; \href{https://github.com/cf3i/Evolving-Multi-Agent-System}{Math}; \href{https://github.com/google-research/google-research/blob/master/mbpp/README.md}{MBPP}; \href{https://github.com/princeton-nlp/tree-of-thought-llm}{Game24}.}
Here HumanEval-Ext., MBPP-Advanced, GPQA-hard100, and GSM8K-Opt. are benchmark-derived profiles, FixBench uses Fixbench-RTL, and workflow Math follows EMAS; where a benchmark appears in multiple settings, each experiment specifies its own cohort, starting agent, and verification contract.

\noindent\textbf{Baselines.}
We benchmark SpecRSI against five groups of baselines:
(1) scientific optimization, following Janus's comparison protocols~\citep{janus2026}: Alg.-Only (AlphaEvolve-style program evolution with a fixed evaluator)~\citep{alphaevolve2025}, Direct-LLM (LAEA-style candidate-quality prediction)~\citep{laea2024}, Embed-GP (a Gaussian-process surrogate on fixed code embeddings)~\citep{snoek2012bayesian}, and Janus;
(2) system-level RSI~\citep{metarsi2026}: Initial, Data-RSI, Harness-RSI, Model-RSI, Fixed D$\to$M$\to$H, a failure-signature Static router, and MetaRSI-v1;
(3) code optimization: DGM~\citep{dgm2025}, SICA~\citep{sica2025}, HGM~\citep{hgm2025}, our non-learning Beam-Spec control with speculative decoding~\citep{leviathan2023}, and FlashEvolve (async/optional)~\citep{flashevolve2026};
(4) prompt optimization: MIPROv2 (with/without demonstrations)~\citep{mipro2024}, Trace~\citep{trace2024}, TextGrad~\citep{textgrad2024}, and GEPA/GEPA+Merge~\citep{gepa2026}; and
(5) workflow optimization: Genesis/Genesis+SOP~\citep{emas2026}, AFlow~\citep{aflow2025}, ADAS~\citep{adas2025}, and EMAS~\citep{emas2026}.
All comparisons match edit permissions, where MetaRSI-H and SpecRSI-H restrict updates to D+H (prompts, tool workflows, and non-test data) with frozen model weights~\citep[Sec.~5.3]{metarsi2026}, and SpecRSI executes direct checks. In cross-backbone transfer tests (Figure~\ref{fig:backbone-performance}), we freeze the shared Qwen-trained improver across all ten scientific and twenty system backbones (nineteen beyond the reference model; see the \href{https://github.com/SpecRSI-GPR/SpecRSI/blob/main/docs/results/model_sources.md}{online reproduction notes}), matching starting agents, contracts, seeds, and evaluation caps ($K$ for Janus/SpecRSI in scientific tasks, and D+H splits and search caps for MetaRSI/SpecRSI in system tasks).

\begin{table}[!htb]\centering

\caption{Scientific and system performance (five-seed mean $\pm$ SD; bold indicates the best comparable mean). (A) Terminal gain (upper) and call-axis AUBC (lower) in native units, following Janus~\citep{janus2026}. (B) Scores (\%) of complete admissible versions under matched D/H/M (upper) or D+H (lower) permissions on Terminal-Bench 2.1 (TB2.1), SWE-Pro, GPQA-hard100 (GPQA), and AIME, following MetaRSI~\citep{metarsi2026}.}
\label{tab:main-benchmarks}\label{tab:main-science}\label{tab:main-system}\label{tab:N11}\label{tab:single-surface}
\begingroup
\fontsize{9}{11}\selectfont
\setlength{\tabcolsep}{0.7pt}\renewcommand{\arraystretch}{1.0}
\renewcommand{\second}[1]{#1}\renewcommand{\best}[1]{\textbf{#1}}
\newcommand{\BCell}[2]{#1\kern.35pt\textpm\kern.35pt#2}
\newcommand{\SciCell}[4]{\shortstack[r]{#1\kern.35pt\textpm\kern.35pt#2\\[1.5pt]#3\kern.35pt\textpm\kern.35pt#4}}
\newcommand{\ScienceTable}{%
\begin{tabularx}{\linewidth}{@{}l*{5}{>{\raggedleft\arraybackslash}X}@{}}\toprule
Method & PyBaMM & AQM & FDTD & Reactor & IRP\\\midrule
Alg.-Only~\citep{alphaevolve2025} & \SciCell{0.12}{0.02}{7.3}{0.1} & \SciCell{0.09}{0.03}{5.1}{0.8} & \SciCell{0.14}{0.05}{8.5}{0.6} & \SciCell{0.11}{0.05}{6.7}{0.2} & \SciCell{0.10}{0.07}{5.8}{0.6}\\
Direct-LLM~\citep{laea2024} & \SciCell{0.21}{0.03}{19.2}{0.8} & \SciCell{0.17}{0.01}{14.3}{0.5} & \SciCell{0.23}{0.02}{22.6}{0.3} & \SciCell{0.18}{0.06}{16.2}{0.1} & \SciCell{0.18}{0.05}{13.7}{0.3}\\
Embed-GP~\citep{snoek2012bayesian} & \SciCell{0.29}{0.03}{30.5}{0.4} & \SciCell{0.26}{0.01}{24.8}{0.1} & \SciCell{0.36}{0.03}{36.2}{0.9} & \SciCell{0.29}{0.03}{27.4}{0.2} & \SciCell{0.28}{0.06}{23.1}{0.1}\\
Janus~\citep{janus2026} & \SciCell{\second{0.35}}{0.05}{\second{42.8}}{0.3} & \SciCell{\second{0.30}}{0.03}{\second{34.7}}{0.9} & \SciCell{\second{0.40}}{0.02}{\second{51.4}}{0.7} & \SciCell{\second{0.33}}{0.02}{\second{37.0}}{0.2} & \SciCell{\second{0.27}}{0.04}{\second{22.6}}{0.1}\\
\rowcolor{OursRow}SpecRSI~(ours) & \SciCell{\best{0.58}}{0.02}{\best{57.7}}{0.4} & \SciCell{\best{0.52}}{0.03}{\best{46.8}}{0.1} & \SciCell{\best{0.64}}{0.01}{\best{67.9}}{0.2} & \SciCell{\best{0.55}}{0.01}{\best{50.3}}{0.4} & \SciCell{\best{0.53}}{0.03}{\best{43.3}}{0.1}\\
\bottomrule\end{tabularx}}

\begin{minipage}[t]{\textwidth}\vspace{0pt}\raggedright

{\centering\CaptionRoman\upshape\mdseries\color{black} A\quad Scientific gain / AUBC\par}\vspace{2pt}
\ScienceTable

\end{minipage}\par\vspace{9pt}
\begin{minipage}[t]{\textwidth}\vspace{0pt}\raggedright

{\centering\CaptionRoman\upshape\mdseries\color{black} B\quad System scores (\%, $\uparrow$)\par}\vspace{2pt}
\begin{tabularx}{\linewidth}{@{}l*{4}{>{\raggedleft\arraybackslash}X}@{}}\toprule
Method & TB2.1 & SWE-Pro & GPQA & AIME\\\midrule
Initial~\citep{metarsi2026} & \BCell{23.6}{0.7} & \BCell{10.2}{0.3} & \BCell{71.2}{1.0} & \BCell{54.9}{0.5}\\
Data-RSI~\citep{metarsi2026} & \BCell{27.4}{0.3} & \BCell{12.4}{0.8} & \BCell{73.8}{0.3} & \BCell{57.6}{0.1}\\
Harness-RSI~\citep{metarsi2026} & \BCell{29.4}{0.6} & \BCell{14.8}{0.3} & \BCell{78.2}{0.7} & \BCell{63.1}{0.8}\\
Model-RSI~\citep{metarsi2026} & \BCell{27.0}{0.3} & \BCell{13.9}{0.1} & \BCell{75.4}{0.6} & \BCell{59.7}{0.2}\\
Fixed D$\to$M$\to$H~\citep{metarsi2026} & \BCell{30.3}{0.6} & \BCell{15.3}{0.3} & \BCell{79.5}{0.2} & \BCell{64.2}{0.6}\\
Static router~\citep{metarsi2026} & \BCell{30.8}{0.2} & \BCell{14.9}{0.3} & \BCell{79.7}{0.9} & \BCell{64.0}{0.2}\\
MetaRSI-v1~\citep{metarsi2026} & \BCell{\second{32.1}}{0.3} & \BCell{\second{19.5}}{0.8} & \BCell{\second{83.8}}{0.2} & \BCell{\second{68.5}}{0.1}\\
\rowcolor{OursRow}SpecRSI~(ours) & \BCell{\best{35.6}}{0.6} & \BCell{\best{23.7}}{0.1} & \BCell{\best{87.3}}{0.3} & \BCell{\best{72.9}}{0.3}\\
\midrule
MetaRSI-H~\citep{metarsi2026} & \BCell{\second{28.6}}{0.5} & \BCell{\second{14.2}}{0.2} & \BCell{\second{77.5}}{0.5} & \BCell{\second{61.7}}{0.2}\\
\rowcolor{OursRow}SpecRSI-H~(ours) & \BCell{\best{32.1}}{0.1} & \BCell{\best{17.3}}{0.5} & \BCell{\best{80.9}}{0.5} & \BCell{\best{65.6}}{0.1}\\

\bottomrule\end{tabularx}

\end{minipage}
\endgroup

\end{table}

\noindent\textbf{Evaluation Metrics.}
We report terminal gain ($\Delta_K$) and area under the budget curve (AUBC) on scientific tasks following Janus~\citep{janus2026}, alongside task scores and mean inference cost ($C$).
Token-accounted searches track initial and remaining budgets $B_0$ and $B_t\ge0$ (in k tokens) separately from the evaluation-call limit $K$, which is fixed across scientific methods at $K=140,130,160,135,120$ for PyBaMM, AQM, FDTD, Reactor, and IRP, respectively. Each call executes one instance's full real-evaluation suite (counting failed runs, but excluding cached and proxy evaluations). Let $s_b$ denote the held-out score of the complete admissible incumbent selected on validation after $b$ calls (with held-out scores never used during search selection), and let $\Delta_b=s_b-s_0$. We compute
\begin{equation}
\Delta_K=s_K-s_0,\quad
\mathrm{AUBC}=\frac12\sum_{b=1}^{K}(\Delta_{b-1}+\Delta_b),\quad
C=\frac{\sum_{i=1}^{M}T_i}{10^3M},
\label{eq:evaluation-metrics}
\end{equation}
where $M$ is the number of evaluated instances and $T_i$ sums all input and output inference tokens over instance $i$'s full execution, including failed or unfinished attempts. Thus $C$ measures online inference cost in k tokens per task (whereas Figure~\ref{fig:motivation}A counts tokens up to the first qualifying outcome or termination, and seed totals or training-token costs are listed separately; Appendix~\ref{app:overall-results}). Scientific gains use dimensionless task scores (Appendix~\ref{app:implementation-details}), and AUBC integrates gain over calls by carrying the final value forward to $K$ after early stopping. Task scores are reported as accuracy or success percentages (with absolute differences in percentage points, pp, and relative gains over the named baseline); higher gain, AUBC, and score and lower cost indicate better performance. Unless stated otherwise, we average three searches per training seed and report mean $\pm$ SD across five seed means.

\noindent\textbf{Implementation Details.}
We implement SpecRSI using Qwen3.5-35B-A3B\footnote{\href{https://huggingface.co/Qwen/Qwen3.5-35B-A3B}{Qwen3.5-35B-A3B}}
as the reference backbone, equipping the Drafter with rank-16 LoRA~\citep{lora2022} and using separate 256-dimensional branches for the Scorer and Policy. Supervised warm-up trains for one epoch using AdamW~\citep{adamw2019} with a learning rate of $2\times10^{-5}$, followed by two reference GVO collection-and-update rounds with $G=3$ independent complete searches per group (see Appendix~\ref{app:implementation-details} for full settings and Appendix~\ref{app:results-learning} for the 20-round extension).

\begin{findingblock}
\begin{figure}[!htbp]
\centering
\includegraphics[width=1.0\columnwidth]{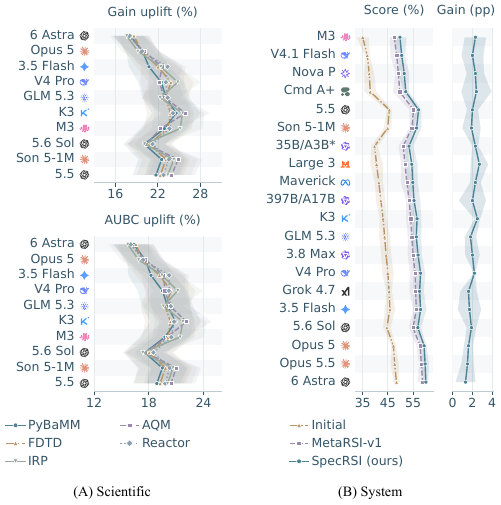}
\caption{Performance across categorical backbones (lines connect five-seed means; shaded bands show $\pm$ SD). (A) Terminal-gain and AUBC uplift over Janus. (B) System scores and seed-paired gains (SpecRSI $-$ MetaRSI-v1). Logos denote model families and axis labels omit family prefixes (Son: Sonnet; P: Premier; Cmd: Command; Qwen3.5: total/active parameters; 3.8 Max: Qwen3.8 Max; Maverick: Llama 4 Maverick; $^*$reference backbone).}
\label{fig:backbone-performance}
\Description{Two side-by-side panels. The left panel shows gain and AUBC uplift over Janus across ten scientific backbones; the right panel shows system scores and seed-paired gains across twenty backbones. All 180 estimates and SDs are retained. Model-family logos sit beside shortened model labels.}
\end{figure}

\subsection{Overall Performance (RQ1)}\label{sec:overall-performance}
In this section, we benchmark SpecRSI across tasks and backbones.
\keypoint{1}~\textbf{Higher scientific gains.} Across the five scientific tasks, SpecRSI improves mean terminal gain and call-axis AUBC over Janus by 60.0\%--96.3\% and 32.1\%--91.6\%, respectively (Table~\ref{tab:main-benchmarks}A).
\keypoint{2}~\textbf{Better system-level successors.} Under matched D/H/M permissions, SpecRSI outperforms MetaRSI-v1 by 3.5--4.4 pp, while restricting updates to D+H retains a 3.48 pp average advantage (Table~\ref{tab:main-benchmarks}B).
\keypoint{3}~\textbf{Gains persist across backbones.} Transferring the frozen improver across models improves all 50 scientific backbone--task pairs over Janus and outperforms MetaRSI-v1 by 1.30--2.70 pp across all twenty system backbones (Figure~\ref{fig:backbone-performance}).
\begin{figure}[!htbp]
\centering
\includegraphics[width=1.0\columnwidth]{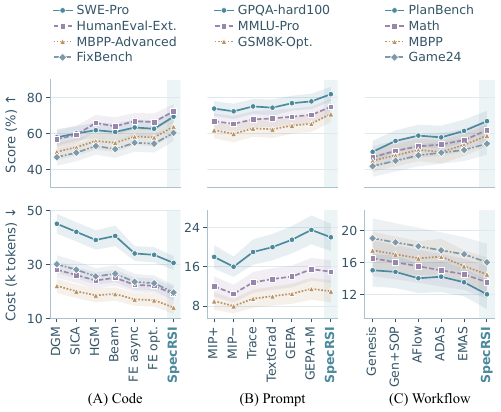}
\caption{Code (A), prompt (B), and workflow (C) performance, showing task score (\%, top row) and mean full-task inference cost (k tokens, bottom row, with separate axis ranges per column). Points and shaded bands show five-seed means $\pm$ SD across connected method categories (Beam: Beam-Spec; FE: FlashEvolve async/optional; MIP$+/-$: MIPROv2 with/without demonstrations; GEPA+M: GEPA+Merge; Gen+SOP: Genesis+SOP).}
\label{fig:surface-heatmaps}
\Description{Six panels arranged in three columns and two rows. Code, prompt, and workflow are the columns; score and full-task inference cost are the rows. Eleven benchmarks are distinguished by color, marker, and line style within each column. Across 146 five-seed estimates, means are marked by points with one-SD shaded bands. All score panels share 30--90 percent axes; the Code, Prompt, and Workflow cost axes cover 10--50, 5.5--27.5, and 9--22.5 thousand tokens, respectively. Methods appear in the same left-to-right order in both rows of each column. SpecRSI is the final method, with its tick label bold and its column lightly shaded.}
\end{figure}

\keypoint{4}~\textbf{Benefits extend across optimization settings.} On the eleven code, prompt, and workflow benchmarks, SpecRSI improves mean scores by 3.5--6.0 pp over the strongest baseline on each task while reducing inference cost by 3.2\%--17.6\% (Figure~\ref{fig:surface-heatmaps}; see Appendix~\ref{app:overall-results} and Figure~\ref{fig:instance-results}(A,B) for instance-level results).
\end{findingblock}

\par\begin{findingblock}
\begin{figure}[!htbp]
\centering
\captionsetup{type=figure}
\includegraphics[width=1.0\columnwidth]{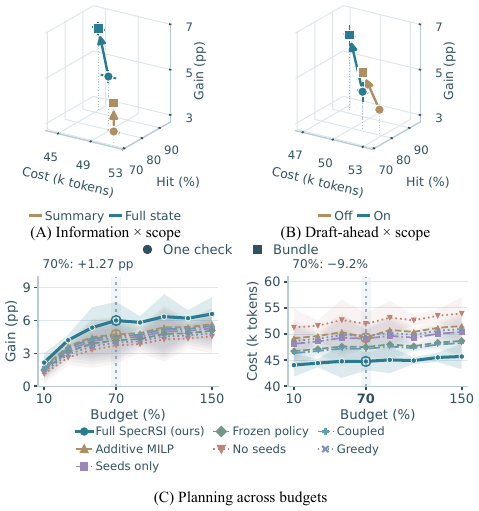}
\caption{Verification scope and planning ablations. (A) Information-scope controls (draft-ahead off) and (B) draft-ahead-scope controls (full-state inputs), reporting accepted gain, search cost (k tokens), and pooled hit rate across $5\times25$ searches per condition. (C) Held-out gain (retaining negative values) and total cost across eight execution-only budget caps (100\% = 65 k tokens, with planning and other overhead included in total cost). Error bars and bands show $\pm$ SD across five seed means; see Table~\ref{tab:key-paired-comparisons} and Appendix~\ref{app:results-planning} for accounting and paired statistics.}
\label{fig:planning-budgets}\label{tab:planning-budgets}\label{tab:N05}\label{tab:N06}
\Description{Panels A and B separately show information-scope and drafting-scope controls, with gain in percentage points, inference-token search cost in thousands of tokens, and hit rate in percent. Arrows connect single to bundled checks. Panel C aligns held-out gain and total cost curves over eight execution-only caps for full SpecRSI and six planning controls; negative gains are retained. Error bars and bands show standard deviations across five seed means; hit rates are pooled over 125 searches per condition.}
\end{figure}

\subsection{VATS Ablations (RQ2)}\label{sec:search-evidence}
In this section, we dissect verification scope and planning in VATS on PyBaMM (Figure~\ref{fig:planning-budgets}).
\keypoint{1}~\textbf{Bundles improve accepted outcomes.}
With full-state inputs and draft-ahead disabled, bundled verification raises accepted gain from 4.8 to 6.3 pp and hit rate (the share of 125 searches returning a complete, admissible agent with positive gain) from 84\% to 92\%, while reducing search cost from 47.96 to 44.99 k tokens.
\keypoint{2}~\textbf{Draft-ahead and bundles yield the best mean outcomes.}
In the $2\times2$ factorial comparison, combining draft-ahead with bundled checks achieves the highest accepted gain (6.02 pp) and hit rate (92\%) alongside the lowest cost (47.64 k tokens).
\keypoint{3}~\textbf{Planning improves the gain--cost trade-off.}
Full SpecRSI achieves the highest mean gain and lowest mean cost across all eight budget caps; at the 70\% cap, it reaches 6.00 versus 4.73 pp for additive MILP planning while using 44.76 versus 49.29 k tokens (9.2\% less).
Under matched inputs, scoring continuation return (CR) also improves gain by 1.32 pp over candidate-quality scoring (CQ) at 7.4\% lower search cost (Figure~\ref{fig:search-learning-controls}A,B; Appendix~\ref{app:results-planning}).

\end{findingblock}

\begin{figure}[!htbp]
\begingroup
\centering
\captionsetup{type=figure,skip=3pt}
\includegraphics[width=1.0\columnwidth]{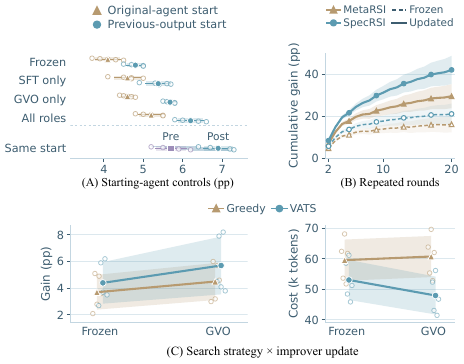}
\caption{Learning the improver (lines and markers show five-seed means with $\pm$ SD shading; open circles in A and C denote individual seeds). (A) Original-agent and previous-output starts alongside the identical-start pre-/post-GVO test. (B) Cumulative gain from previous-output starts after round-1 initialization, comparing Frozen improvers with Updated improvers trained each round. (C) VATS$\times$GVO factorial control, where total cost sums training and eight evaluation searches per seed (protocols in Appendix~\ref{app:results-learning}).}
\label{fig:improver-learning}\label{fig:rounds-main}
\Description{Three panels show learning effectiveness across five seeds. Panel A compares Frozen, SFT-only, GVO-only, and all-role updates starting from the original task agent or the previous output, plus a paired comparison from an identical starting agent. Panel B shows cumulative gains across rounds 2 through 20: before-training improvers stay frozen, while after-training improvers update each round with GVO for SpecRSI and MetaRSI's own rule. Panel C is a Greedy-versus-VATS by frozen-versus-GVO factorial experiment: accepted gains are 3.70, 4.50, 4.40 and 5.70 percentage points; per-seed total costs including training are 59.60, 60.80, 53.10 and 48.00 thousand tokens, respectively. Each seed's total cost sums training and eight independent evaluation searches. Filled marks show means, shaded intervals show standard deviations, and open circles in A and C show individual seed values.}
\endgroup
\end{figure}

\subsection{Learning Effectiveness (RQ3)}\label{sec:improver-rounds}
In this section, we evaluate improver learning across starting agents and multi-round trajectories (Figure~\ref{fig:improver-learning}).
\keypoint{1}~\textbf{GVO updates improve search quality.}
When starting searches either from the original task agent or from the previous round's output (without resetting the improver), GVO-only updates raise gains from 4.10/4.80 to 4.60/5.68 pp, respectively (Figure~\ref{fig:improver-learning}A). In the matched factorial comparison (Figure~\ref{fig:improver-learning}C), GVO adds 0.80 pp under Greedy search and 1.30 pp under VATS; overall, full SpecRSI outperforms Greedy with a frozen improver by 2.00 pp at 19.5\% lower total token cost (including training).
\keypoint{2}~\textbf{Learning helps from the same starting agent.}
Holding the starting task agent fixed in a paired test, GVO updates increase gain from 5.7 to 6.9 pp (Figure~\ref{fig:improver-learning}A), yielding a paired improvement of 1.20 pp (95\% CI $[1.112,1.288]$, positive across all five seeds; Table~\ref{tab:key-paired-comparisons}).
\keypoint{3}~\textbf{Gains accumulate across rounds.}
Across 20 rounds (comparing frozen before-training improvers against after-training improvers updated each round via GVO or MetaRSI's rule), cumulative gain reaches 42.1 pp for updated SpecRSI versus 29.6 pp for updated MetaRSI and 21.2 pp for frozen SpecRSI, with positive mean per-round gains throughout rounds 2--20 (Figure~\ref{fig:improver-learning}B; see also frozen-improver transfer in Figure~\ref{fig:backbone-performance}B and Appendix~\ref{app:results-learning}).

\FloatBarrier
\begingroup
\setlength{\intextsep}{3pt}
\begin{table}[!htb]
\centering

\caption{Resource trade-offs (five-seed mean $\pm$ SD) cover candidate-plan limits (A), full-run lifecycle costs (B), and parallel execution (C): 8/20 workers run $6.7/12.9\times$ faster at $+19.0/55.2\%$ worker time vs.\ 1 (Appendix~\ref{app:results-resources}).}
\label{tab:resources}
\captionsetup{font={small,captionroman},skip=3pt}

\fontsize{8.5}{10}\selectfont
\setlength{\tabcolsep}{0.7pt}
\renewcommand{\arraystretch}{1.0}
\begin{minipage}[t]{0.46\linewidth}
\vspace{0pt}
{\centering\CaptionRoman\upshape\mdseries\color{black} (A) Candidate-plan limit\par}\vspace{2pt}
\begin{tabular*}{\linewidth}{@{\extracolsep{\fill}}lrr@{}}
\toprule
Cap & Gain (pp) & \shortstack{Planning cost\\(k tokens)}\\
\midrule
8 & $5.2\!\pm\!0.35$ & $1.8\!\pm\!0.12$\\
16 & $5.8\!\pm\!0.22$ & $2.7\!\pm\!0.15$\\
32 & $6.3\!\pm\!0.27$ & $3.9\!\pm\!0.21$\\
64 & $6.5\!\pm\!0.16$ & $5.6\!\pm\!0.23$\\
\bottomrule
\end{tabular*}

\par\vspace{3pt}
{\centering\CaptionRoman\upshape\mdseries\color{black} (B) Lifecycle cost (k tokens)\par}\vspace{2pt}
\begin{tabular*}{\linewidth}{@{\extracolsep{\fill}}lrr@{}}
\toprule
Category & MetaRSI-v1 & SpecRSI\\
\midrule
Acquisition & $14.5\!\pm\!0.41$ & $14.2\!\pm\!0.35$\\
Training & $38.6\!\pm\!0.96$ & $34.8\!\pm\!0.87$\\
Search & $37.2\!\pm\!0.96$ & $36.2\!\pm\!0.86$\\
Reuse checks & $5.1\!\pm\!0.26$ & $4.8\!\pm\!0.26$\\
Copies & $6.3\!\pm\!0.26$ & $6.1\!\pm\!0.25$\\
Failed work & $6.8\!\pm\!0.40$ & $6.3\!\pm\!0.32$\\
Planning & $4.2\!\pm\!0.20$ & $3.9\!\pm\!0.20$\\
\midrule
Total & $112.7\!\pm\!3.44$ & \cellcolor{OursRow}$\mathbf{106.3}\!\pm\!3.09$\\
\bottomrule
\end{tabular*}

\end{minipage}\hfill
\begin{minipage}[t]{0.52\linewidth}
\vspace{0pt}
{\centering\CaptionRoman\upshape\mdseries\color{black} (C) Parallel execution\par}\vspace{2pt}
\fontsize{8}{10}\selectfont\setlength{\tabcolsep}{1.2pt}
\begin{tabular*}{\linewidth}{@{\extracolsep{\fill}}rrrr@{}}
\toprule
Wkr. & \shortstack{Gain\\(pp)} & \shortstack{Time\\(s)} & \shortstack{Total worker\\time (worker-s)}\\
\midrule
\textbf{1} & $6.30\!\pm\!0.20$ & $1287\!\pm\!21.79$ & $1287.00\!\pm\!21.79$\\[0.90712pt]
2 & $6.40\!\pm\!0.18$ & $672\!\pm\!10.91$ & $1340.00\!\pm\!24.51$\\[0.90712pt]
3 & $6.40\!\pm\!0.26$ & $458\!\pm\!9.80$ & $1372.00\!\pm\!28.25$\\[0.90712pt]
4 & $6.40\!\pm\!0.20$ & $354\!\pm\!6.40$ & $1412.00\!\pm\!26.08$\\[0.90712pt]
5 & $6.40\!\pm\!0.14$ & $291\!\pm\!5.43$ & $1451.00\!\pm\!28.08$\\[0.90712pt]
6 & $6.50\!\pm\!0.21$ & $248\!\pm\!5.43$ & $1484.00\!\pm\!33.52$\\[0.90712pt]
7 & $6.50\!\pm\!0.19$ & $216\!\pm\!4.47$ & $1508.00\!\pm\!31.76$\\[0.90712pt]
\textbf{8} & $6.50\!\pm\!0.23$ & $192\!\pm\!4.47$ & $1531.40\!\pm\!34.51$\\[0.90712pt]
9 & $6.50\!\pm\!0.19$ & $174\!\pm\!4.47$ & $1560.90\!\pm\!39.00$\\[0.90712pt]
10 & $6.50\!\pm\!0.20$ & $160\!\pm\!4.47$ & $1594.84\!\pm\!44.63$\\[0.90712pt]
12 & $6.60\!\pm\!0.21$ & $139\!\pm\!4.47$ & $1663.50\!\pm\!54.46$\\[0.90712pt]
14 & $6.60\!\pm\!0.24$ & $125\!\pm\!4.47$ & $1743.24\!\pm\!63.23$\\[0.90712pt]
16 & $6.60\!\pm\!0.18$ & $114\!\pm\!4.85$ & $1818.30\!\pm\!75.62$\\[0.90712pt]
18 & $6.64\!\pm\!0.23$ & $106\!\pm\!4.47$ & $1904.56\!\pm\!80.43$\\[0.90712pt]
\textbf{20} & $6.70\!\pm\!0.23$ & $100\!\pm\!4.47$ & $1996.80\!\pm\!89.01$\\[0.90712pt]
\bottomrule
\end{tabular*}

\end{minipage}
\par\vspace{2pt}

\end{table}
\endgroup

\subsection{Resource Trade-offs (RQ4)}\label{sec:resource-tradeoffs}
In this section, we examine candidate-plan limits, lifecycle costs, and parallel execution (Table~\ref{tab:resources}).
\keypoint{1}~\textbf{More plans yield diminishing gains.}
Raising the candidate-plan limit from 8 to 32 increases gain from 5.2 to 6.3 pp, whereas doubling it to 64 adds only 0.2 pp while increasing planning cost from 3.9 to 5.6 k tokens (Table~\ref{tab:resources}A).
\keypoint{2}~\textbf{SpecRSI uses fewer lifecycle tokens.}
SpecRSI achieves lower mean token usage across all seven lifecycle categories, reducing full-run expenditure (including training and serialized copies) from 112.7 k tokens for MetaRSI-v1 to 106.3 k tokens ($-5.7\%$; Table~\ref{tab:resources}B).
\keypoint{3}~\textbf{Parallel workers reduce latency but use more worker time.}
Scaling from one to eight concurrent workers cuts elapsed wall-clock time from 1,287 to 192 s ($6.7\times$) at the cost of 19.0\% more total worker time, while twenty workers further reduce latency to 100 s ($12.9\times$) with 55.2\% more worker time as mean gain rises from 6.3 to 6.7 pp (Table~\ref{tab:resources}C).
\FloatBarrier
\section{Conclusion}\label{sec:conclusion}
In this work, we present SpecRSI, whose two-level loop extends incompletely evaluated parents but adopts
only complete, admissible versions.
VATS allocates edits and complementary checks by predicted verified continuation
return; GVO updates the improver from completed searches under a fixed
verification contract.
SpecRSI improves terminal-gain and AUBC means over Janus in all 50 scientific backbone--task pairs
and exceeds MetaRSI-v1 by 1.30--2.70 pp in score across
twenty backbones.
Full planning outperforms six planning controls in mean held-out gain and
mean total token cost across all eight budget caps.
Starting-artifact controls support gains from improver updates beyond
retaining stronger task agents.
Instance-level comparisons show higher PyBaMM gain and SWE-Pro success with lower token cost.
These results show that planning verification and learning the improver jointly improve agent performance and token efficiency.

\FloatBarrier\label{main:end}
\endgroup
\bibliographystyle{content/references_style}
\bibliography{content/references_compact}
\appendix
\setcounter{figure}{0}\setcounter{table}{0}\setcounter{algorithm}{0}
\renewcommand{\thefigure}{S\arabic{figure}}
\renewcommand{\thetable}{S\arabic{table}}
\renewcommand{\thealgorithm}{S\arabic{algorithm}}
\renewcommand{\theHfigure}{supp.\arabic{figure}}
\renewcommand{\theHtable}{supp.\arabic{table}}
\clearpage
\enableappendixcontents\listofappendices
\clearpage
\raggedbottom
\renewcommand{\topfraction}{.95}
\renewcommand{\textfraction}{.08}
\section{Extended Related Work}\label{app:extended-related}\par\noindent\textbf{Recursive self-improvement.}
Recent agent-optimization frameworks---including DGM, SICA, ADAS, AFlow, and GEPA---iteratively refine agent programs, workflows, or prompts~\citep{dgm2025,sica2025,adas2025,aflow2025,gepa2026}.
To guide multi-step evolution, HGM expands lineages using descendant outcomes while scheduling evaluations separately~\citep{hgm2025}, Hyperagents directly modifies the meta-agent~\citep{hyperagents2026}, MetaRSI adapts improvement operators and their scheduling~\citep{metarsi2026}, and Janus co-evolves target programs with screening proxies under real-validation gates~\citep{janus2026}.
By contrast, SpecRSI learns both successor drafting and verification allocation directly from the net verified returns of complete searches.

\par\noindent\textbf{Speculative decoding.}
Standard speculative decoding accelerates generation by verifying cheap draft tokens in parallel with a target model~\citep{leviathan2023}.
In agentic settings, FlashEvolve asynchronously overlaps evolution stages~\citep{flashevolve2026}.
SpecRSI instead brings speculation to persistent agent revisions: \textbf{only adopted versions require complete verification}, whereas intermediate ancestors may remain partially evaluated.

\par\noindent\textbf{Learning for agent improvement.}
Whereas GRPO estimates group-relative advantages across responses to the same prompt~\citep{grpo2024} and MAS$^2$ trains meta-agents via Collaborative Tree Optimization~\citep{mas22025}, GVO assigns credit across independent complete searches launched from matched starting states.
Its search-level returns reward verified gains net of all incurred costs---including preparation as well as failed or unfinished branches---using realized suffix returns to train the Scorer and whole-search advantages to update the Drafter and Policy.

\tcbset{appendixsettings/.style={enhanced,
  colback=black!2,colframe=black!65,colbacktitle=black!74,coltitle=white,
  boxrule=0.45pt,arc=1mm,boxsep=0pt,left=1.55pt,right=1.55pt,
  top=0.3pt,bottom=0.3pt,toptitle=1.2pt,bottomtitle=1.2pt,
  fonttitle=\footnotesize\bfseries,
  before skip=0pt,after skip=0pt,
  grow to left by=2pt,grow to right by=2pt}}

\section{Implementation Details}\label{app:reproduction}
\label{app:implementation-details}\label{app:additional-reproduction}
\label{app:aqm-reference}\label{app:cost-definitions}\label{app:results-resources}\label{app:parallel-workers}
In this section, we elaborate on the architecture, optimization, benchmark, and accounting settings used in Section~\ref{sec:experimental-setup} and Algorithms~\ref{alg:main}--\ref{alg:epoch}, summarized in Table~\ref{tab:implementation-settings}.

\par\Needspace{7\baselineskip}\begingroup
\setlength{\parskip}{0pt}
\setlength{\smallskipamount}{1pt}
\captionsetup{skip=3pt}
\captionof{table}{Implementation and evaluation settings.}
\label{tab:implementation-settings}
\footnotesize
\setlength{\tabcolsep}{2pt}
\renewcommand{\arraystretch}{1.0}
\begin{tcolorbox}[appendixsettings,breakable,width=\linewidth,enlarge left by=0pt,enlarge right by=0pt,title={Implementation and evaluation}]
\raggedright
\textbf{Architecture.} Frozen BF16 text representations; separate Scorer/Policy branches with typed relations and four-head set attention~\citep{settransformer2019}. Inputs: observed outcomes, missing checks, remaining resources, proposed work.\par
\textbf{Warm-up.} Component minibatch 1; Drafter/Policy/Scorer targets: executed legal edits/legal choices/realized returns. Unexecuted alternatives receive no targets.\par
\textbf{VATS.} SciPy \href{https://docs.scipy.org/doc/scipy/reference/generated/scipy.optimize.milp.html}{\texttt{milp}}/HiGHS~\citep{highs2018}. Training/default limits: targets, pair unions, singleton closures $2/2/2$ / $4/4/4$; uncached Scorer queries, retained plans, accepted extensions/seed $(q_{\max},K_\Omega,s_{\max})=(4,8,4)$ / $(96,32,4)$.\par
\textbf{GVO.} $\lambda=\mu=0$: verified gain under hard caps, with cost features and feasibility active (nonzero penalties use a separate profile). $\le1$ joint SGD step/collection; Scorer/actor rates $5\!\times\!10^{-4}/10^{-3}$; $(\epsilon_c,\beta_{\rm pair},\beta_{\rm KL})=(0.2,1,0.02)$. Zero momentum/weight decay; no gradient-norm clipping.\par
\textbf{Execution.} Training: 2,048-token sequences, one H20/scientific domain. Inference: 32,768-token limit, separate H20. Other backbones: frozen transfer via provider APIs.\par
\par\smallskip\hrule\smallskip
\textbf{Scientific evaluation: 12 held-out instances per task.} PyBaMM v23.5; $J=1-\mathrm{MSE}/\mathrm{MSE}_{\rm ref}$ (dimensionless), $\mathrm{MSE}=N^{-1}\sum_i(V_{\rm pred,i}-V_{\rm true,i})^2$ ($\mathrm V^2$). Reference: fixed unoptimized industry defaults, distinct from root. Acceptance: physical bounds and convergence.\par
\textbf{AQM.} 12 parameters/12 scenarios; $U=0.4g_{\rm raw}/g_{\max}-0.3d_{95}/d_{95,\rm base}-0.2\ell-0.1(1-f)$. Throughput/link capacity, 95th-percentile delay/initial-controller delay; $\ell$: loss fraction; $f$: Jain fairness. Legal parameters, stable queues, no sustained packet loss.\par
\textbf{Other science.} Scores normalized to $[0,1]$. FDTD: convergence, crosstalk $\le-20$ dB. Reactor: convergence, safe temperature/pressure. IRP: feasible constraints, service level $\ge95\%$.\par
\textbf{Public AQM.} ns-3.44, seven RED/ECN parameters; three TCP NewReno patterns $\times$ two seeds/case, 8 s/run; 20-Mb/s dumbbell, four flows. Mean utility $g-d_{95}-2\ell-0.5(1-f)$; $g$: normalized goodput; $d_{95}$: 95th-percentile end-to-end delay (s). Legal parameters and all six outputs required; measured simulator time charged.\par
\textbf{System tasks.} Terminal-Bench 2.1, SWE-Pro v1.0, GPQA-hard100, AIME 2025 I/II (30 problems).\par
\par\smallskip\hrule\smallskip
\textbf{Accounting.} Table~\ref{tab:resources}: all-round totals/seed, including fine-tuning input/output tokens and serialized snapshots/trajectories. Defaults: PyBaMM, 65-k-token search budget, 32 candidate plans, one H20; worker threads share global state/caches; worker time sums active execution. Figure~\ref{fig:search-learning-controls}E: forward input/output and forced-continuation tokens; search charged separately, backward computation in GPU time.\par

\end{tcolorbox}
\endgroup\par

\section{More Experimental Results}\label{app:extended-experiments}
In this section, we present instance-level RQ1 evaluations, additional RQ2--RQ3 controls, and qualitative case studies on PyBaMM and SWE-Pro, complementing Sections~\ref{sec:overall-performance}--\ref{sec:improver-rounds}.
\subsection{Instance-level Performance}\label{app:overall-results}
\label{app:experiments}\label{app:additional}\label{app:coding-results}
\begin{figure}[!htbp]
\centering
\includegraphics[width=1.0\linewidth]{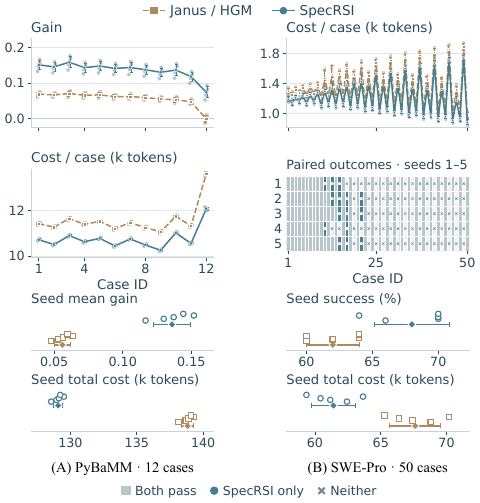}
\caption{Instance-level outcomes and cost (in k tokens) across five seeds on (A) 12 categorical PyBaMM cases (120 records) and (B) 50 SWE-Pro cases (500 records; 155 both-pass, 15 SpecRSI-only, 0 HGM-only, and 80 neither-pass pairs). Small points denote individual seed runs, upper lines and bars show per-case means $\pm$ SD, and lower panels summarize per-seed mean gain/success and seed-total cost ($\pm$ SD).}
\label{fig:instance-results}\label{fig:instance-pybamm}\label{fig:instance-swepro}
\Description{Two side-by-side panels at one-column width show all 620 method-seed-case observations. PyBaMM shows gain and cost across 12 categorical cases; SWE-Pro shows cost across 50 cases and all 250 paired binary outcomes. Four lower plots retain all five seed summaries and SDs. Baselines Janus/HGM use tan squares and dashed curves; SpecRSI uses teal circles and solid curves. The paired matrix uses neutral cells for both pass, teal circles for SpecRSI-only, and crosses for neither; HGM-only count is zero.}
\end{figure}

In this subsection, we extend RQ1 to paired instances on PyBaMM and SWE-Pro (Figure~\ref{fig:instance-results}) under the protocols in Table~\ref{tab:additional-protocols}.
\keypoint{1}~\textbf{Higher gains accompany lower cost.} SpecRSI improves PyBaMM terminal gain by 142.6\% over Janus and SWE-Pro resolution rate by 6.0 pp over HGM, while reducing token expenditure by 7.0\% and 9.2\%, respectively.

\begin{table}[htbp]

\centering
\caption{Paired comparisons for RQ1 (upper) and RQ2/RQ3 (lower). All seven differences are positive in all five seed pairs, using SpecRSI minus baseline or the named direction. Units are pp except native-unit PyBaMM gain/AUBC ($\dagger$held-out; last two rows: in-search gains).}
\label{tab:key-paired-comparisons}
\small

\setlength{\tabcolsep}{2pt}\renewcommand{\arraystretch}{1.05}
\begin{tabularx}{\linewidth}{@{}Xrr@{}}\toprule
Comparison & Difference & 95\% CI\\\midrule
PyBaMM gain: vs. Janus$^\dagger$ & $+0.07992$ & $[0.07057,0.08926]$\\
SWE-Pro success: vs. HGM$^\dagger$ & $+6.00$ & $[4.244,7.756]$\\
Reference-system mean: vs. MetaRSI-v1$^\dagger$ & $+3.90$ & $[3.834,3.966]$\\
PyBaMM AUBC: vs. Janus$^\dagger$ & $+14.9$ & $[14.776,15.024]$\\
\midrule
Full $-$ Additive MILP (70\%)$^\dagger$ & $+1.27$ & $[1.118,1.422]$\\
Full state: bundle $-$ single & $+1.50$ & $[1.475,1.525]$\\
Same start: updated $-$ original & $+1.20$ & $[1.112,1.288]$\\
\bottomrule\end{tabularx}

\end{table}

\subsection{Search and Learning Controls}\label{app:results-planning}\label{app:evidence-ablations}\label{app:fixed-information}
\label{app:results-learning}\label{app:supervision-controls}\label{app:scorer-configurations}
\begin{figure}[!htbp]
\centering
\includegraphics[width=1.0\columnwidth]{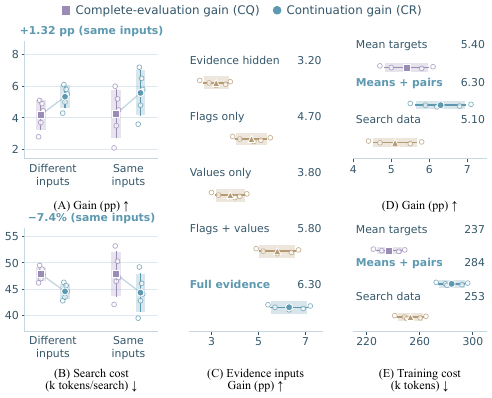}
\caption{Scoring, evidence, and supervision controls (five-seed means $\pm$ SD shading, with open circles showing individual seeds and costs in k tokens). (A,B) CQ versus CR gain and search-inference cost; (C) evidence-input ablations; (D,E) held-out supervision gain and training cost (protocols in Appendix~\ref{app:results-planning}).}
\label{fig:search-learning-controls}
\Description{Five panels compare scoring targets, test-time controller masks, and training supervision. Matched continuation-gain targets raise mean accepted gain by 1.32 percentage points and reduce search cost by 7.4 percent. Evidence conditions retain resource inputs and complete Solver and verification information; the observed-values condition retains check values while masking availability flags and structural relations. Full evidence has mean gain 6.30 percentage points. On identical acquired plans, mean-return targets give 5.40 plus or minus 0.57 percentage points and 237.00 plus or minus 9.53 thousand training tokens; adding return-difference targets gives 6.30 plus or minus 0.65 and 284.00 plus or minus 9.71. Ordinary-search data give 5.10 plus or minus 0.57 and 253.00 plus or minus 9.45, with acquisition expense matched within two percent. Shading shows standard deviations across five seeds; open circles show individual values.}
\end{figure}

In this subsection, we present additional VATS and GVO controls in Figure~\ref{fig:search-learning-controls}, with protocols detailed in Table~\ref{tab:additional-protocols}.
\keypoint{1}~\textbf{Continuation targets and evidence improve search.} Under matched inputs, continuation-return scoring (CR) improves gain by 1.32 pp (95\% CI $[1.184,1.456]$) over complete-evaluation gain scoring (CQ) while using 3.52 k fewer tokens (7.4\%), and full evidence yields 6.30 pp versus 3.20--5.80 pp under restricted inputs.
\keypoint{2}~\textbf{Supervision and GVO improve accepted gain.} On identical plans, adding pairwise targets raises gain from 5.40 to 6.30 pp at higher training cost; relative to Greedy/frozen, VATS and GVO add 0.70 and 0.80 pp individually and 2.00 pp jointly (95\% paired CI $[0.452,3.548]$, interaction $+0.50$ pp, 95\% CI $[-0.520,1.520]$; Figure~\ref{fig:improver-learning}C).

\par\Needspace{7\baselineskip}\begingroup
\captionof{table}{Additional evaluation protocols (costs in k tokens; supervision summaries report five-seed mean $\pm$ SD).}
\label{tab:additional-protocols}
\footnotesize
\setlength{\tabcolsep}{2pt}
\renewcommand{\arraystretch}{1.0}
\begin{tcolorbox}[appendixsettings,breakable,title={Evaluation protocols}]
\raggedright
\textbf{Instance comparisons.} Matched starts/token caps/cases; five paired within-seed means, pointwise 95\% intervals $\bar d\pm t_{0.975,4}s_d/\sqrt5$. System scores weight four tasks equally. Held-out gain follows Equation~\eqref{eq:evaluation-metrics}, retaining negative values.\par
\textbf{Figure~\ref{fig:motivation}.} First complete admissible output with $[J(c)-J(r)]/|J(r)|\ge0.05$ (absolute gain $\ge0.05$ if $J(r)=0$), or best admissible output at termination. Per-task thresholds; equal task weights; token accounting: Section~\ref{sec:experimental-setup}.\par
\par\smallskip\hrule\smallskip
\textbf{Cohorts.} Figure~\ref{fig:planning-budgets} uses separate cohorts for A, B, and C; Figure~\ref{fig:search-learning-controls} uses separate cohorts for A--B, C, and D--E.\par
\textbf{Figure~\ref{fig:search-learning-controls}A,B.} PyBaMM, original/matched: 25/8 searches/seed. Original: CQ omits resource inputs and uses its own stopping rule; CR uses full inputs. Matched: identical network/state inputs, stopping/ACCEPT, training data, and 65-k-token budget; only targets differ.\par
\textbf{Figure~\ref{fig:search-learning-controls}C.} Eight PyBaMM searches/seed, 65 k tokens; same trained Scorer/Policy masked at test time, with complete Solver inputs and real $\Gamma$ verification intact. Flags/observations/relations: available/missing; values incl.\ admissibility/validity; dependency/shared-setup/source-reuse. Hidden/flags-only/values-only/restored/full: $(000)/(100)/(010)/(110)/(111)$. Remaining budget/attempts always visible.\par
\textbf{Figures~\ref{fig:planning-budgets}A,B / \ref{fig:improver-learning}C.} Mean final ACCEPT gain; failures/root returns count as zero in-search gain. Figure~\ref{fig:planning-budgets}C accounting follows Appendix~\ref{app:execution}; at 10\%, mean total $44.05=6.32$ drafting/verification $+2.08$ initialization/contract loading $+35.65$ planning/Scorer queries/state checks.\par
\textbf{Figure~\ref{fig:search-learning-controls}D,E.} Same snapshots/trajectories: mean returns with/without pairwise return-difference targets. Ordinary-search: mean targets, acquisition cost matched within 2\%. Gain: independent held-out complete searches. Mean-only / +pair / ordinary gain: $5.40\pm0.57$ / $6.30\pm0.65$ / $5.10\pm0.57$ pp; training cost: $237.00\pm9.53$ / $284.00\pm9.71$ / $253.00\pm9.45$.\par
\par\smallskip\hrule\smallskip
\textbf{Figure~\ref{fig:improver-learning}A.} Shared $\Theta_0$: Frozen/GVO-only/All roles update none/Drafter--Policy/all. SFT-only: random adapters/heads, one epoch on full search trajectories, actor cross-entropy + Scorer regression; warm-up uses fewer demonstrations.\par
\textbf{Figure~\ref{fig:improver-learning}B.} Each curve: 20 rounds $\times$ three searches; updated curves apply one GVO step or native MetaRSI update per round, while frozen curves retain initial improvers. All curves carry the previous best complete admissible output (or unchanged root), reset evidence/budgets, and sum within-seed gains over rounds 2--20.\par
\textbf{Figure~\ref{fig:improver-learning}C.} PyBaMM; matched starts/permissions/contracts/budgets. Greedy scores current candidate gain without continuation scoring, draft-ahead, or bundles. Final frozen improver: eight independent searches/seed after training. Total cost follows the figure caption, including training collections.\par

\end{tcolorbox}
\endgroup\par

\definecolor{CaseSetup}{HTML}{2F65B0}
\definecolor{CaseBlocked}{HTML}{B43C3C}
\definecolor{CaseSuccess}{HTML}{21804A}
\definecolor{CaseAnalysis}{HTML}{B35C12}
\FloatBarrier
\subsection{Qualitative Case Studies}\label{app:qualitative-cases}
In this subsection, we present two representative cases in Tables~\ref{tab:case-pybamm} and~\ref{tab:case-metarsi}, separating single-seed execution traces from five-seed summaries. Search costs count input/output and tool-return tokens, excluding simulator or test runtime and separately tracked LoRA training (the \href{https://github.com/SpecRSI-GPR/SpecRSI/blob/main/docs/results/online_supplement.md\#table-o2-dh-csvcut-case}{online supplement} provides an additional D+H \texttt{csvcut} case with frozen model weights).

\begin{table}[htbp]
\centering
\caption{PyBaMM calibration case (the median gain-difference case among the 12 instances in Figure~\ref{fig:instance-results}A). Scores and gains are dimensionless; costs are in k tokens.}
\label{tab:case-pybamm}
\footnotesize
\setlength{\tabcolsep}{3pt}\renewcommand{\arraystretch}{1.02}
\begin{tabularx}{\linewidth}{@{}X@{}}\toprule
\textbf{Setup.} \textcolor{CaseSetup}{PyBaMM v23.5; shared root and 140 complete-call budget. Calibrates seven NMC--graphite single-particle-model parameters against a 1C discharge curve at $25^\circ$C, including diffusivity, SEI resistance, and exchange-current values/SOC dependence. The root uses constant diffusivity/SEI resistance, has maximal mid-SOC voltage error, underestimates low-SOC resistance, and has a 100-fold exchange-current prefactor error.}\\\midrule
\textbf{Score.} Appendix~\ref{app:implementation-details} defines scoring and admissibility. Fixed-reference RMSE: $\simeq33.8$ mV; root: $J(r)=0.280$, RMSE $28.7$ mV. The 15-mV target ($J\simeq0.80$) is not an admissibility threshold.\\\midrule
\textbf{Janus trace.} Calibrates constant diffusivity and then SEI resistance, simulating each step; \textcolor{CaseBlocked}{proxy screening rejects the joint edit after five proxy evaluations and two full 800-s simulations.}\\
\textbf{SpecRSI trace.} \textcolor{CaseSuccess}{Drafts piecewise SOC-dependent diffusivity, exponential SOC-dependent SEI resistance, and exchange-current/concentration corrections; one bound check and one final simulation verify all three edits without intermediate full simulations.}\\\midrule
\textbf{Janus, five seeds.} Scores: $0.339,0.344,0.347,0.353,0.349$; mean $\mathbf{0.3464}$ (RMSE equivalent $27.3$ mV), gain $\mathbf{0.0664}$, cost $\mathbf{11.264}$. Token shares: simulation interaction/proxy/generation $=62/21/17\%$.\\
\textbf{SpecRSI, five seeds.} Scores: $0.409,0.421,0.427,0.440,0.433$; mean \textcolor{CaseSuccess}{$\mathbf{0.4260}$} (RMSE equivalent $25.6$ mV), gain \textcolor{CaseSuccess}{$\mathbf{0.1460}$}, cost \textcolor{CaseSuccess}{$\mathbf{10.504}$}. Token shares: simulation interaction/bounds/generation/planning $=39/8/36/17\%$.\\\midrule
\textcolor{CaseAnalysis}{\textbf{Analysis.} By composing three sequential drafts before full simulation, SpecRSI achieves 119.9\% higher mean gain at 6.7\% lower token cost.}\\\bottomrule
\end{tabularx}
\end{table}

\begin{table}[htbp]
\centering
\caption{Full D/H/M \texttt{csvstat} case (Table~\ref{tab:main-benchmarks}B). Search costs are five-seed means in k tokens.}
\label{tab:case-metarsi}
\footnotesize
\setlength{\tabcolsep}{3pt}\renewcommand{\arraystretch}{1.02}
\begin{tabularx}{\linewidth}{@{}X@{}}\toprule
\textbf{Setup.} \textcolor{CaseSetup}{Shared root and 3.5-k-token budget under data, harness, and task-model editing. Repairs accumulation and adds \texttt{--weight} for weighted mean/variance, requiring 28 target tests plus clean syntax/lint. The root passes 21/28 (75\%), lacks weighting, and incurs $10^{-6}$-order error on mixed large/small values.}\\\midrule
\textbf{MetaRSI-v1 trace.} Replaces summation with Kahan compensation and then attempts weighted means; \textcolor{CaseBlocked}{two boundary-test failures halt that branch after three full \texttt{pytest} runs, leaving the final version at \textbf{26/28} (92.9\%) with weighting incomplete.}\\
\textbf{SpecRSI trace.} \textcolor{CaseSuccess}{Drafts compensated summation, weight parsing, weighted mean/variance, and boundary handling alongside a task-model LoRA update; syntax, precision, weighting-subset, and lightweight post-update checks precede a single full \texttt{pytest}/\texttt{pyflakes} run, passing \textbf{28/28}.}\\\midrule
\textbf{Search cost.} MetaRSI-v1: \textbf{3.185}; SpecRSI: \textcolor{CaseSuccess}{\textbf{2.974} ($-6.6\%$)}. Full-test interaction/generation/other search inference account for $69/22/9\%$ and $31/38/17\%$, respectively, with SpecRSI spending 14\% on staged checks.\\\midrule
\textcolor{CaseAnalysis}{\textbf{Analysis.} SpecRSI completes both features with fewer full test runs and lower token cost, jointly verifying code and adapter updates before adoption.}\\\bottomrule
\end{tabularx}
\end{table}

\section{Algorithms and Training}
\label{app:methods}\label{app:rsi-objects}\label{app:contract}\label{app:plans}
\label{app:learning}\label{app:execution}
In this section, we state the execution, MILP, and training rules supplementing Section~\ref{sec:method} and Algorithms~\ref{alg:main}--\ref{alg:epoch} (with hyperparameters in Appendix~\ref{app:implementation-details}).

\begingroup
\makeatletter
\def\@parfont{\bfseries\upshape}
\newcommand{\AlgLine}[1]{\State\parbox[t]{\dimexpr\linewidth-\ALG@thistlm-1pt\relax}{\raggedright #1}}
\makeatother

\begin{algorithm}[htbp]
\fontsize{8}{9}\selectfont
\caption{SpecRSI: Frozen Search Collection and Learning}\label{alg:main}\label{alg:compact}
\begin{algorithmic}[1]
\Require Admissible $c_0$; warm-up $\Theta_0$; $\Gamma,\nu$; $B_0,N_0$, workers; $G\ge2$, $K_{\rm out}$; carry/reset rule.
\For{$k=0,\ldots,K_{\rm out}-1$}
\AlgLine{$r\gets c_k$; $\bar\phi_k\gets\phi_k$; freeze $\Theta_k$.}
\For{$g=1,\ldots,G$ independently in collection mode}
\AlgLine{\AlgAnchor{inner:start}Initialize $S_0^-$ at $r$ with root evidence/budgets; $\mathcal D\gets\varnothing$, $t\gets0$, $S_{\rm end}\gets S_0^-$.}
\While{$\bar o_B\le B_t^-$, preparation caps permit, and $N_t>0$}
\AlgLine{$(S_t,\Omega_t,\omega_t,o_t,\ell_t^m,\mathcal P_t)\gets\textbf{VATS}(S_t^-)$ (Alg.~\ref{alg:prepare}); \AlgAnchor{resume:selection-return}log paid preparation once; $S_{\rm end}\gets S_t$.}
\AlgLine{\AlgAnchor{resume:main}If $\omega_t=\Accept$, set $S_{t+1}^-=S_t$; log zero-cost work, increment $t$, and \textbf{break}.}
\AlgLine{Execute DRAFT~\eqref{eq:draft} or VERIFY via $\Acquire_\Gamma$; retain $Z_t$.}
\AlgLine{Update tree/valid evidence, debit work and attempts, and log $S_{t+1}^-$; $S_{\rm end}\gets S_{t+1}^-$; $t\gets t+1$.}
\EndWhile
\AlgLine{\AlgAnchor{inner:end}Store $\Accept(S_{\rm end})$~\eqref{eq:accept}, terminal state/index, costs, and $\mathcal D_k^{(g)}\gets\mathcal D$.}
\EndFor
\AlgLine{$(\Theta_{k+1},\text{ledger})\gets\textbf{GVO}(\Theta_k,\{\mathcal D_k^{(g)}\}_{g=1}^G)$ (Alg.~\ref{alg:epoch}); retain IDs and ledger.}
\AlgLine{Reset $c_{k+1}\gets c_0$, or carry an admissible output by the registered selector.}
\EndFor
\AlgLine{\Return $\Theta_{K_{\rm out}}$, outputs, and logs.}
\end{algorithmic}
\end{algorithm}

\paragraph{KL regularization}\label{app:kl-regularization}
Writing $p_{g,j}^{\vartheta}=p_\vartheta(\cdot\mid H_{g,j},\mathcal Y_{g,j})$, we define
\begin{equation}
\mathcal L_{\rm KL}=\frac1G\sum_{g=1}^G\sum_{j=1}^{J_g}\sum_{y\in\mathcal Y_{g,j}}p_{g,j}^{\vartheta}(y)\log\frac{p_{g,j}^{\vartheta}(y)}{p_{g,j}^{\vartheta_0}(y)},\label{eq:kl}
\end{equation}
where the post-warm-up reference $\vartheta_0$ remains fixed, detached, and strictly positive on the shared legal support, with both distributions evaluated under identical logged contexts, temperatures, masks, and exploration transforms.

\begin{algorithm}[htbp]
\fontsize{8}{9}\selectfont
\caption{Verification-Aware Tree Search (VATS)}\label{alg:prepare}
\begin{algorithmic}[1]
\Require $S^-$; frozen $\bar\phi,\psi,\nu$; mode; caps $\bar o,q_{\max},s_{\max},K_\Omega$.
\AlgLine{$\mathcal R\gets\Remaining(S^-)$; $\widetilde S\gets S^-[B\gets B^-\! -\bar o_B]$; initialize cache, $q\gets0$, and preparation meter.}
\AlgLine{Select targets under $\nu$; form $\mathcal M_0$ from validated MILP covers and feasible singleton closures.}
\AlgLine{$\mathcal M\gets\mathcal M_0$; define $W(A)=\widehat Q_{\bar\phi}(\widetilde S,\Verify(A))$ only for returned queries.}
\For{each $A_0\in\mathcal M_0$ in registered order}
\AlgLine{$A\gets A_0$; $s\gets0$; query/cache $W(A)$ if permitted, metering attempted cache misses in $q,o$.}
\While{$s<s_{\max}$, $q<q_{\max}$, and preparation fits}
\AlgLine{Query a capped subset of feasible $U=A\cup D_j$; meter every attempted cache miss in $q$ and expense.}
\AlgLine{If no available $W(U)>W(A)$, \textbf{break}; else take the best improvement, set $A\gets U$, add $A$ to $\mathcal M$, increment $s$.}
\EndWhile
\EndFor
\AlgLine{Build/cap $\Omega$ from ACCEPT, draft plans and $\Verify(A)$ ($A\in\mathcal M$); validate, deduplicate and retain required alternatives.}
\AlgLine{Settle $o\le\bar o$ and $o_B\le\bar o_B$ including Policy; $S\gets\Debit(S^-,o_B)$; log total preparation $o$.}
\AlgLine{Choose $\omega$ by~\eqref{eq:actor}: sample in collection, maximize logits in deployment; finalize elapsed time $\ell^m$.}
\AlgLine{\Return $S,\Omega,\omega,o,\ell^m,\mathcal P$, logging choice probability, context/support, mode, snapshots and preparation.}
\end{algorithmic}
\end{algorithm}

\paragraph{Execution}
Deployment runs a single frozen search, and forced continuations trigger no outer-loop updates. Disjoint preparation and execution ledgers charge all failed or unfinished work without imputing missing records as zero labels.
At each step, the menu retains $\Accept$, feasible seeds, available edits, singleton checks, and affordable completions regardless of their isolated scores; every non-$\Accept$ action consumes at least one attempt, leaving $\Accept$ as the sole option when no other work is feasible, and inconsistent caps are rejected.
Preparation tracks actual and reserved budget charges $o_B,\bar o_B$ alongside total expense and cap $o,\bar o$ ($0\le o_B\le\bar o_B\le B^-$); in Figure~\ref{fig:planning-budgets}C, setting $o_B=\bar o_B=0$ means $B$ caps drafting and verification while separately capped preparation enters total cost, with cache hits requiring matched state, scorer snapshot, and plan.

\begin{algorithm}[htbp]
\fontsize{8}{9}\selectfont
\caption{Group Verification Optimization (GVO)}\label{alg:epoch}
\begin{algorithmic}[1]
\Require Frozen $\Theta_k$; $G\ge2$ complete logs; saved states/IDs; fixed post-warm-up $\vartheta_0$; update settings.
\AlgLine{Compute complete-search $R_g,A_g$ by~\eqref{eq:reward}.}
\AlgLine{$\mathcal D_{\rm actor}\gets\bigcup_g\mathcal D_g^{\rm sampled}$; $\mathcal D_Q\gets\bigcup_g\mathcal D_g^{\rm plan\ starts}$ with labels~\eqref{eq:valuefit}; $\mathcal D_{\rm pair}\gets\varnothing$.}
\For{prespecified paid $S$, legal $a\in\mathcal W_S$, $s=1,\ldots,n_a$}
\AlgLine{Clone paid $S$, root/environment; log acquisition separately; start suffix accounting.}
\AlgLine{Force $a$ at Alg.~\ref{alg:main}'s work dispatch; skip initialization/first preparation, then continue with frozen $\Theta_k$ to termination.}
\AlgLine{Add logged starts/returns~\eqref{eq:valuefit} to $\mathcal D_Q$, preserving IDs/missingness/charges; $\mathcal D_{\rm actor}$ unchanged.}
\EndFor
\AlgLine{Form matched-state $\mathcal D_{\rm pair}$ with $\widehat\delta_{ab}=\overline G_a-\overline G_b$~\eqref{eq:pair}.}
\AlgLine{Evaluate all gradients at $(\vartheta_k,\phi_k)$; empty losses are zero; detach behavior probabilities and $A_g$.}
\AlgLine{If Scorer records exist: $\phi\gets\phi_k-\eta_\phi\nabla_\phi(\mathcal L_Q+\beta_{\rm pair}\mathcal L_{\rm pair})$; otherwise $\phi\gets\phi_k$.}
\AlgLine{If sampled choices exist: $\vartheta\gets\vartheta_k-\eta_\vartheta\nabla_\vartheta(\mathcal L_{\rm GVO}+\beta_{\rm KL}\mathcal L_{\rm KL})$ by~\eqref{eq:gvo},\eqref{eq:kl}; otherwise $\vartheta\gets\vartheta_k$.}
\AlgLine{$(\theta,\psi)\gets\vartheta$; $\Theta_{k+1}\gets(\theta,\psi,\phi,\nu)$; $\bar\phi_{k+1}\gets\phi$; collect fresh searches next round.}
\AlgLine{\Return $\Theta_{k+1}$ and ledger; select checkpoints on development data only.}
\end{algorithmic}
\end{algorithm}

\par\noindent\begin{minipage}{\columnwidth}
\paragraph{Completion MILP}
Each job $j$ specifies covered keys $K_j$, reserved expense $\bar c_j$, attempts $\bar n_j$, and a setup group $\gamma(j)$ with cost $s_\gamma$. Using binary decision variables $x_j,y_\gamma$ for jobs and setups under target set $K\ne\varnothing$ and planning budget $B_{\rm plan}=B^- -\bar o_B$, the Solver solves
\begin{equation}\label{eq:completion-milp}
\begin{gathered}
\min_{x_j,y_\gamma\in\{0,1\}} C=\sum_j\bar c_jx_j+\sum_\gamma s_\gamma y_\gamma,\\
\sum_{j:\xi\in K_j}x_j\ge1\ (\xi\in K),\quad C\le B_{\rm plan},\quad 1\le\sum_j\bar n_jx_j\le N,\\
x_j\le x_i\ (i\text{ prerequisite of }j),\quad x_i+x_j\le1\ (i,j\text{ conflict}),\\
x_j\le y_{\gamma(j)},\qquad y_\gamma\le\sum_{j:\gamma(j)=\gamma}x_j .
\end{gathered}
\end{equation}
Only independently validated feasible covers seed the search (valid timeout incumbents need not be optimal); our implementation uses SciPy/HiGHS with round-robin targets and an independently seeded reservoir prior to observing outcomes.
\end{minipage}\par

\endgroup

\end{document}